\documentclass[10pt,twocolumn,letterpaper]{article}
\usepackage{amsmath,amsfonts}
\usepackage{algorithmic}
\usepackage{algorithm}
\usepackage{array}
\usepackage[caption=false,font=normalsize,labelfont=sf,textfont=sf]{subfig}
\usepackage{textcomp}
\usepackage{stfloats}
\usepackage{verbatim}
\usepackage{graphicx}
\usepackage{multirow}
\usepackage{algorithm}
\usepackage{algorithmic}
\usepackage{bm}
\usepackage{color}

\usepackage{colortbl}
\usepackage{makecell}
\usepackage{amsmath}
 \usepackage[pagenumbers]{cvpr} % To force page numbers, e.g. for an arXiv version

\definecolor{cvprblue}{rgb}{0.21,0.49,0.74}
\usepackage[pagebackref,breaklinks,colorlinks,allcolors=cvprblue]{hyperref}

\def\paperID{*****} % *** Enter the Paper ID here
\def\confName{CVPR}
\def\confYear{2025}

\title{DEAL: Data-Efficient Adversarial Learning for High-Quality Infrared Imaging}

\author{
	Zhu Liu$^{\dag}$, Zijun Wang$^{\dag}$, Jinyuan Liu$^\ddag$, Fanqi Meng$^\dag$, Long Ma$^\dag$, Risheng Liu$^{\dag}$\thanks{Corresponding author.}\\	
\normalsize$^\dag$School of Software Technology, Dalian University of Technology\\
\normalsize $^\ddag$School of Mechanical Engineering, Dalian University of Technology\\
{\tt \small  liuzhu@mail.dlut.edu.cn,wzijun6@gmail.com, rsliu@dlut.edu.cn}
}

\begin{document}
\maketitle
\begin{abstract}
	Thermal imaging is often compromised by dynamic, complex degradations caused by hardware limitations and unpredictable environmental factors. The scarcity of high-quality infrared data, coupled with the challenges of dynamic, intricate degradations, makes it difficult to recover   details using existing methods. In this paper, we introduce thermal degradation simulation integrated into the training process via a mini-max optimization, by modeling these degraded factors as adversarial attacks on thermal images. The simulation is dynamic to maximize objective functions, thus capturing a broad spectrum of degraded data distributions. This approach enables training with limited data, thereby improving model performance.
	Additionally, we introduce a dual-interaction network that combines the benefits of spiking neural networks with scale transformation to capture degraded features with sharp spike signal intensities. This architecture ensures compact model parameters while preserving efficient feature representation. Extensive experiments demonstrate that our method not only achieves superior visual quality under diverse single and composited degradation, but also delivers a significant reduction in processing when trained on only fifty clear images, outperforming existing techniques in efficiency and accuracy. The source code  will be available at \url{https://github.com/LiuZhu-CV/DEAL}.
\end{abstract}

%\begin{figure*}[thb]
%	\centering
%	\includegraphics[width=0.98\textwidth]{./pic/first}
%	
%	\caption{{ 
		%%					We first illustrate the thermal-specific degradation in the left subFig.~(a), aiming to simultaneously address the complex and dynamic infrared degradations uniformly. "LR", "LC", and "SN" denote low resolution, low contrast, and stripe noise, respectively. The zoomed-in regions in subFig.~(b) demonstrate the performance of our method under individual degradations, coupled corruptions, and various semantic perception tasks, which indicates significant superiority over state-of-the-art methods.
		%			}}
%
%	\label{fig:first}
%\end{figure*}

\section{Introduction}
Infrared imaging~\cite{wang2023revelation,bao2023heat}, as a robust and insensitive technique for capturing thermal radiation with salient features, plays a significant role in various visual perception tasks, such as multi-modality image fusion~\cite{liu2020bilevel,liu2021searching}, small target detection~\cite{li2024contourlet}, and semantic segmentation~\cite{liu2023multi,liu2023bi}. Due to its ability to effectively capture and convey temperature information, infrared imagery demonstrates reliability and environmental adaptability in adverse conditions~\cite{ma2022toward}. 
\begin{figure*}[thb]
	\centering
	\includegraphics[width=0.99\textwidth]{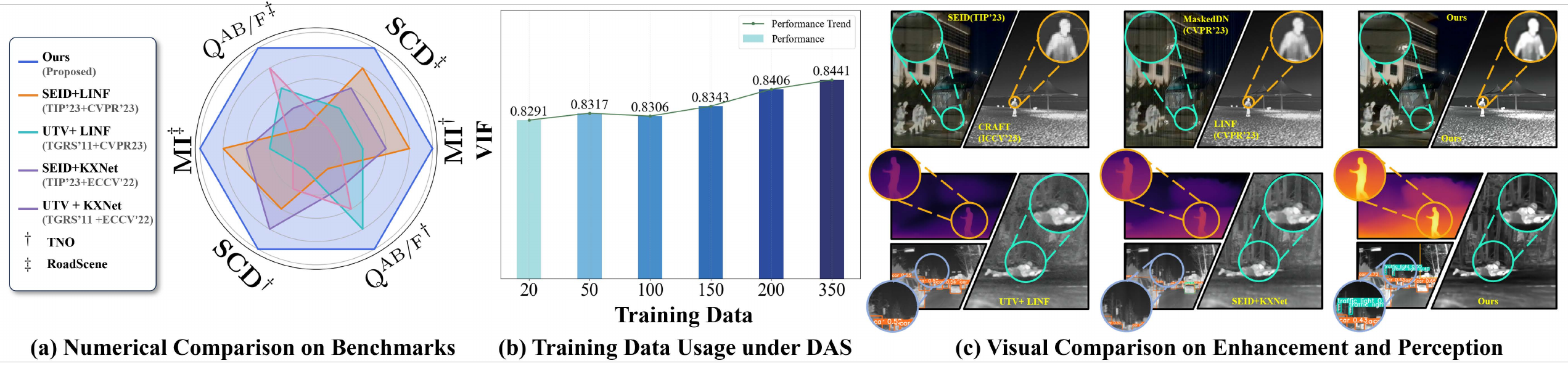}
	\vspace{-0.5em}
	\caption{{Comprehensive evaluation: numerical benchmark comparisons, training data efficiency of DAS, and perceptual comparisons.	
	}}
	
	\label{fig:first}
\end{figure*}

Unfortunately, infrared modality is inevitably characterized by complex degradation factors in real-world dynamic and extreme imaging conditions~\cite{liu2024infrared,liu2022twin}. These corruptions are naturally detrimental to visual human observation and downstream semantic perception tasks. 
Recently, numerous deep learning-based methods have been proposed, which can be roughly divided into two typical methodologies: multi-modality image fusion and single-image infrared enhancement. Various attempts for image fusion have been made to design information measurements (e.g., salience measuring~\cite{liu2024task},  feature richness~\cite{xu2020u2fusion})  cross-modal attention~\cite{liu2023multi,wang2022unsupervised}, and correlation-driven fusion~\cite{liu2024coconet}), thereby mitigating degradation through  visible features. However, in extreme scenes,  degradation of visible light is severe. 

Specifically, in the realm of single-image infrared enhancement~\cite{song2023simultaneous,cai2024exploring,song2023fixed}, learning-based methods targeting specific single degradations are widely investigated, having achieved promising breakthroughs. For instance, Generative Adversarial Networks (GANs)~\cite{liu2024searching} have garnered significant attention for infrared super-resolution, incorporating recursive attention network~\cite{liu2021infrared}, degradation model-guided learning~\cite{chen2024modeling}, and more. Introducing supplementary information (e.g., visible modality), multi-modality fusion modules~\cite{huang2022infrared} and transfer learning strategies~\cite{huang2021infrared} have been proposed for image super-resolution. Moreover, multi-scale feature representations are widely utilized for infrared stripe denoising~\cite{xu2022single}. 
However, two major obstacles hinder the advancement of infrared enhancement research. (1) Scarcity of high-quality thermal data: Infrared imaging is affected by dynamic changes due to temperature-driven thermal radiation and inherent hardware limitations. Selecting appropriate models that can address various degradation factors and intensities is challenging~\cite{jiang2024multispectral,liu2018learning,wang2023interactively}. (2) Lack of effective mechanisms tailored for thermal degradation: Existing methods mostly depend on models designed for visible imaging, which are not optimized for unique thermal characteristics. Our objective is to offer a comprehensive solution for complex degraded scenes, effectively utilizing limited training data.

To partially mitigate these issues, we present the first attempt at establishing a unified paradigm for various thermal degradation. We formulate the learning process as an adversarial framework, separating it into dynamic degradation generation and image enhancement. The adversarial degradation aims to challenge the enhancement network by applying gradient ascent on the training objective, adaptively simulating degraded imaging with diverse levels of thermal-specific corruptions, including stripe noise, low contrast, and spatial resolution. Meanwhile, the enhancement process focuses on restoring the textural details of these degradation factors. We introduce a dynamic adversarial training strategy for progressive degradation generation and infrared image enhancement. Additionally, to capture unique infrared degradation characteristics, we propose a dual-interaction network combining paired scale transformation operations and a spiking-inspired module with dense connections.
To summarize, our contributions are four-fold:

\begin{itemize}
	\item We make the attempt to address various degradations simultaneously in infrared images. Introducing dynamic generation makes our method more applicable to the limitation of high-quality training data.
	\item From a learning standpoint, we design a dynamic adversarial solution, which effectively strengthens the performance of the enhancement network.
	\item From the architectural side, we introduce a dual interaction network to better capture degradation, fully exploring the potential of spiking characteristics with nimble parameters and efficient inference.
	\item Extensive experiments conducted on various degraded scenarios including single and composited degradation factors and perception tasks  such as depth estimation and detection substantiate the significant advancements of proposed method, using only fifty images for training.
\end{itemize}

\begin{figure*}[thb]
	\centering
	\includegraphics[width=0.98\textwidth]{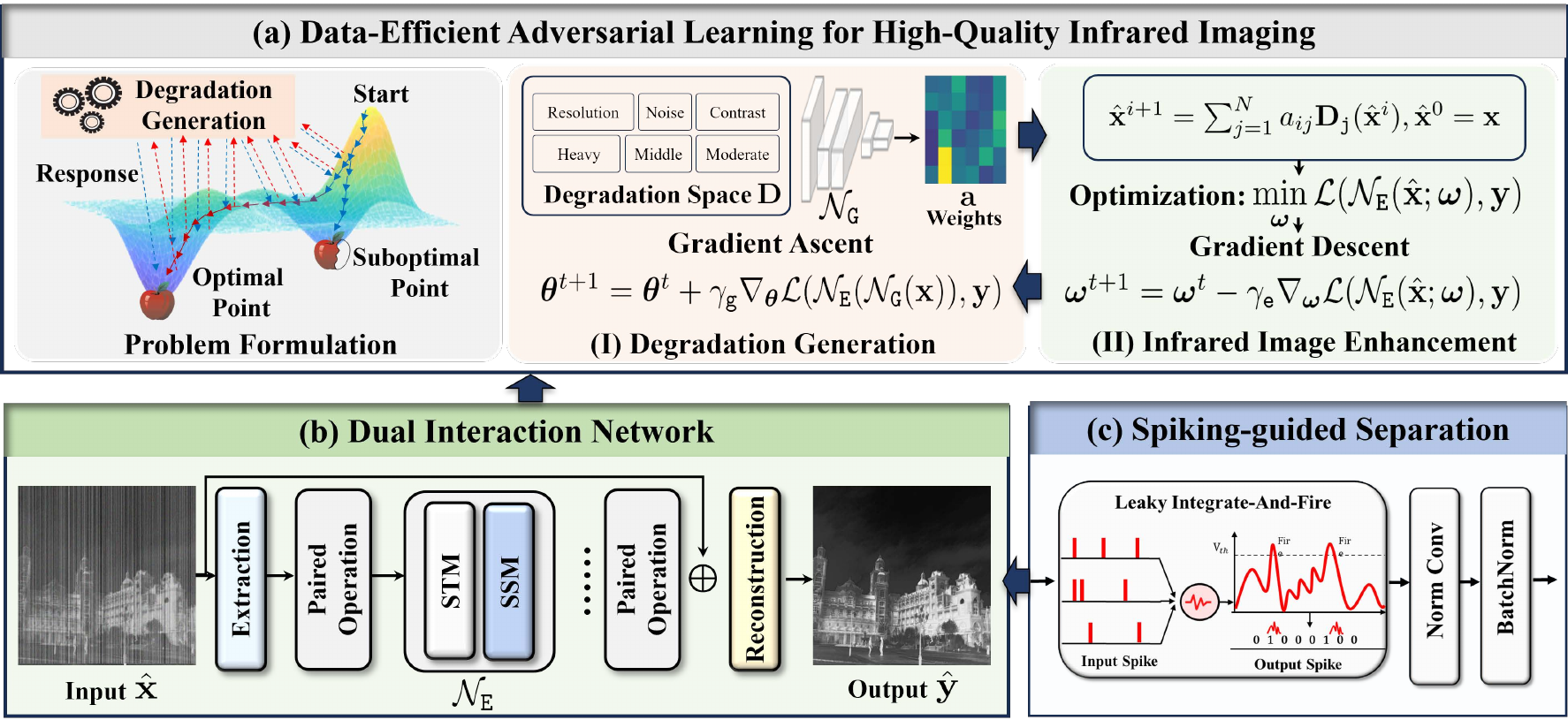}
	\vspace{-0.5em}
	\caption{{ 
			Schematic of the major components of proposed paradigm. We  present 
			a data-efficient adversarial learning strategy, which constructs the dynamic degradation generation to guide the image enhancement procedure with a Dynamic Adversarial Solution (DAS) at  (a). 
			The concrete architecture of dual interaction network, consisting of Scale Transform Module (STM) and Spiking-guided Separation Module (SSM) is shown at (b). Spiking-guided Separation to capture sharp intensities of thermal degradation is depicted at  (c).
	}}
	
	\label{fig:illustration}
\end{figure*}

\section{Proposed Method}
%In this part, we  elaborately detail the proposed degradation-aware learning strategy. We first present the problem formulation to construct an adversarial relationship between degradation generation and infrared image formulation.  Last we propose a hierarchical adversarial solution to address this optimization formulation.

% 红外作为一种恶劣条件下进行工作的模态 多模态互补 很重要， 但红外本身也有一些限制

%现有的解决方案只解决了单一降质情形 缺乏对不同降质的灵活适应能力

%红外降质纹理缺失，大多方法无法建立长程依赖，

%因此提出两种手段 从建模，求解与结构上进行解决
\subsection{Problem Formulation}
In real-world thermal imaging, multiple degradations may occur simultaneously, resulting in a scarcity of high-quality training data. This situation also presents a challenge, as it requires the selection of various learning models to address different types and intensities of corruption.  Dynamically generating various thermal-specific degradations to train an all-in-one model is crucial  for infrared imaging.

\noindent\textbf{Adversarial formulation of degradation.}
To establish the relationship between thermal-specific degradation generation and image enhancement, we adhere to the  adversarial learning paradigm to enhance the robustness of enhancement and alleviate the amount of training data. A hierarchical mini-max optimization is defined as follows:

\begin{align}
	&	\min\limits_{\bm{\omega}}   \mathcal{L}(\mathcal{N}_\mathtt{E}(\hat{\mathbf{x}};\bm{\omega}),\mathbf{y}),\label{eq:main2}\\
	&	\mbox{ s.t. } \left\{
	\begin{aligned}
		\hat{\mathbf{x}} &= \mathcal{N}_\mathtt{G}(\mathbf{x};\bm{\theta}^{*}), \\
		\bm{\theta}^{*} &= \arg\max\limits_{\bm{\theta}} \mathcal{L}(\mathcal{N}_\mathtt{E}(\mathcal{N}_\mathtt{G}(\mathbf{x};\bm{\theta});\bm{\omega}),\mathbf{y}),
	\end{aligned}
	\right.	
	\label{eq:constraint2}
\end{align}
where  $\mathcal{N}_\mathtt{G}$ and $\mathcal{N}_\mathtt{E}$ represent the degradation generation with parameters $\bm{\theta}$ and the  enhancement network with parameter $\bm{\omega}$. $\mathbf{x}$, $\hat{\mathbf{x}}$, and $\mathbf{y}$ denote the original infrared images, corrupted inputs, and the corresponding ground truths, respectively. $\mathcal{L}$ denotes the loss functions of image enhancement. The lower-level maximization sub-problem aims at generating changeable combinations of various thermal degradation, which can significantly impair the performance of the image enhancement network. Thus, the lower-level objective is to maximize losses related to infrared enhancement. The upper-level sub-problem involves a general optimization procedure for image enhancement based on the training of generated degradation $\hat{\mathbf{x}}$, aiming to restore images from various degradation conditions with high robustness. Additionally, mini-max adversarial optimization can equip the enhancement network with a complex degradation-aware capability. 
Additionally, the pipeline of entire framework is depicted in Fig.~\ref{fig:illustration} (a). 

\begin{algorithm}[htb] 
	\caption{Dynamic Adversarial  Solution (DAS).}\label{alg:framework}
	\begin{algorithmic}[1] 
		\REQUIRE Infrared datasets with $\{\mathbf{x},\mathbf{y}\}$,  learning rate $\gamma_\mathtt{e}$,$\gamma_\mathtt{g}$ and other necessary hyper-parameters.
		%		\STATE Preparing  pairs $\{\mathbf{x},\mathbf{y}\}$ with .
		%		\STATE \% \emph{Warm start of $\mathcal{N}\circ\mathcal{T}$.} 
		
		\STATE \% \emph{Warm starting the enhancement network.} 
		
		\STATE $	\bm{\omega}^{t+1} = \bm{\omega}^{t} -\gamma_\mathtt{e}\nabla_{\bm{\omega}} \mathcal{L}(\mathcal{N}_\mathtt{E}(\hat{\mathbf{x}};\bm{\omega}),\mathbf{y})$;
		
		\WHILE {not converged}
		
		\STATE \% \emph{Optimization of infrared image enhancement.} 
		\STATE Generating the degraded samples by $\mathcal{N}_\mathtt{G}$ and setting batch as 
		$\{ \hat{\mathbf{x}}_{1}, \mathbf{y}_{1}, \cdots \hat{\mathbf{x}}_{N}, \mathbf{y}_{N} \}$;
		
		\STATE $\bm{\omega}^{t+1} = \bm{\omega}^{t} -\gamma_\mathtt{e}\nabla_{\bm{\omega}} \mathcal{L}(\mathcal{N}_\mathtt{E}(\hat{\mathbf{x}};\bm{\omega}),\mathbf{y})$;
		
		\STATE \% \emph{Gradient update of corruption generation.}
		\STATE $\bm{\theta}^{t+1} = \bm{\theta}^{t} + \gamma_\mathtt{g}\nabla_{\bm{\theta}} \mathcal{L}(\mathcal{N}_\mathtt{E}(\mathcal{N}_\mathtt{G}(\mathbf{x};\bm{\theta});\bm{\omega}),\mathbf{y})$;
		\ENDWHILE
		
		\RETURN  $\bm{\omega}^{*}$.
	\end{algorithmic}
\end{algorithm}

Initially, with regard to the degradation generation network $\mathcal{N}_\mathtt{G}$, we incorporate a classifier~\cite{liu2022image} to generate weights that simulate the degradation process. This classifier takes the original infrared image $\mathbf{x}$ as input and produces a differentiable weight matrix $\mathbf{a}$ to control the various degrees of various degradation. Denoting one type of degradation as $\mathbf{D}$, the  generation can be formulated as:
\begin{equation}
	\hat{\mathbf{x}}^{i+1} =\sum_{j=1}^{N} {a}_{ij} \mathbf{D}_\mathtt{j}(\hat{\mathbf{x}}^{i}), \hat{\mathbf{x}}^{0} =\mathbf{x}.
	\label{eq:degradation}
\end{equation}

We consider three major degradations in thermal imaging, including stripe streaks, low-resolution, and low contrast. In detail, infrared striped noise typically arises due to non-uniformities in the sensor, such as the effects of diverse column responses of the focal plane array. The low resolution of infrared imaging is widely acknowledged, caused by the propagation characteristics of thermal radiation. Limited by sensor performance, temperature differences, and signal processing algorithms, low contrast in infrared images can stem from various factors. 
Furthermore, for the network of infrared image enhancement $\mathcal{N}_\mathtt{E}$, we propose a dual interaction network to uniformly process degradation.  The detailed architecture is proposed in Section~\ref{sec:din}.

\noindent\textbf{Dynamic adversarial  solution.}  We propose a dynamic adversarial solution for the training procedure, providing an optimization strategy for both  Eq.~\eqref{eq:main2} and Eq.~\eqref{eq:constraint2}.

In detail, directly utilizing adversarial corruptions can easily result in the non-convergence of enhancement and failure in scene restoration. Therefore, we initially warm-start the enhancement network $\mathcal{N}_\mathtt{E}$, which is frozen the degradations generated by $\mathcal{N}_\mathtt{G}$ for several epochs based on gradient descent techniques, which can be formulated as:
\begin{equation}
	\bm{\omega}^{t+1} = \bm{\omega}^{t} -\gamma_\mathtt{e}\nabla_{\bm{\omega}} \mathcal{L}(\mathcal{N}_\mathtt{E}(\hat{\mathbf{x}};\bm{\omega}),\mathbf{y}).
	\label{eq:warmstart}
\end{equation}

Inspired by hierarchical training strategies (e.g., generative adversarial learning~\cite{huang2021infrared} and adversarial attack~\cite{liu2023paif}), we construct a single-loop alternative solution. Since the adoption of  relaxed aggregation  with learnable weights $\mathbf{a}$, we can optimize the generation parameters $\bm{\theta}$ through gradient ascent, which can be written as:
\begin{equation}
	\bm{\theta}^{t+1} = \bm{\theta}^{t} + \gamma_\mathtt{g}\nabla_{\bm{\theta}} \mathcal{L}(\mathcal{N}_\mathtt{E}(\mathcal{N}_\mathtt{G}(\mathbf{x};\bm{\theta});\bm{\omega}),\mathbf{y}).
	\label{eq:att}
\end{equation}

After obtaining the degraded images $\hat{\mathbf{x}}$, we further utilize gradient descent with Eq.\eqref{eq:warmstart} to update $\bm{\omega}$ for the enhancement network. We construct a single-loop optimization~\cite{liu2024moreau,liu2023optimization,liu2019convergence} to update the generation and enhancement network alternately. The   solution is summarized in Alg.~\ref{alg:framework}.

\noindent\textbf{Discussion with  augmentation techniques.} Existing  techniques (\textit{e.g.,} Augmix~\cite{hendrycks2019augmix}, AdaAugment~\cite{yang2024adaaugment}) typically rely on static, predefined operations, such as rotation, translation, or adjusting brightness for the whole dataset, lacking adaptability to the dynamic and complex degradations. These techniques do not dynamically adjust the type and intensity of degradation based on the model’s weaknesses during training, thus falling short in enhancing model performance.  In contrast, our method introduces adversarial degradation generation, treating degradation as an adaptive attack against the enhancement network.  

\begin{table*}[tb]
	\centering
	\footnotesize
	\renewcommand{\arraystretch}{1.1}
	
	\setlength{\tabcolsep}{4mm}{
		\begin{tabular}{|c|c|c|c|c|c|c|c|c|c|}
			\hline
			Levels                    & Metrics & MaskedDN & GF     & SEID   & WDNN   & WFAF   & UTV    & LRSID  & Ours            \\ \hline
			\multirow{5}{*}{\begin{tabular}[c]{@{}c@{}}Moderate\end{tabular}}    & MI     & 3.0901 & 3.2738 & 3.2665 & 3.3383 & 3.3204 & \cellcolor{red!15}\textbf{3.4257} & 3.1504 &\cellcolor{blue!10} 3.3972 \\ \cline{2-10} 
			& VIF    & 0.8714 & 0.9217 & \cellcolor{blue!10} 0.9446 & 0.9252 & 0.9130 & 0.9302 & 0.8270 & \cellcolor{red!15}\textbf{0.9607}  \\ \cline{2-10} 
			& SD     & 10.668 & 10.668 & 10.638 & 10.647 & 10.669 &\cellcolor{blue!10} 10.690 & 10.655 & \cellcolor{red!15}\textbf{11.050} \\ \cline{2-10} 
			&  $\mathrm{Q^{AB/F}}$     & 0.4539 & 0.5001 & \cellcolor{blue!10}0.5017 & 0.4937 & 0.4861 & \cellcolor{red!15}\textbf{0.5157} & 0.4468 & 0.4523 \\ \cline{2-10} 
			& SCD    & 1.3668 & 1.3757 & \cellcolor{blue!10}1.4278 & 1.3400 & 1.3450 & 1.3002 & 1.3509 &\cellcolor{red!15} \textbf{1.4467}  \\ \hline
			\multirow{5}{*}{\begin{tabular}[c]{@{}c@{}}Heavy\end{tabular}} & MI     & 2.6092 & 3.0228 & 3.0241 & 3.1129 & 3.0752 & 3.0752 &\cellcolor{blue!10} 3.1374 &\cellcolor{red!15}\textbf{3.2443} \\ \cline{2-10} 
			& VIF    & 0.9581 & 1.0483 & 1.0721 & 1.0685 & 1.0508 &\cellcolor{blue!10} 1.0854 & 1.0681 &\cellcolor{red!15}\textbf{1.0982} \\ \cline{2-10} 
			& SD     & 10.277 & 10.288 & \cellcolor{blue!10} 10.327 & 10.283 & 10.285 & 10.294 & 10.280 &\cellcolor{red!15} \textbf{10.575} \\ \cline{2-10} 
			&  $\mathrm{Q^{AB/F}}$      & 0.3731 & 0.5003 & 0.5039 & 0.4953 & 0.4870 & \cellcolor{blue!10} 0.5183 & 0.5152 &\cellcolor{red!15} \textbf{0.5520}  \\ \cline{2-10} 
			& SCD    & 1.4596 & 1.4619 & \cellcolor{blue!10} 1.4882 & 1.4198 & 1.4238 & 1.4009 & 1.4204 &\cellcolor{red!15} {\textbf{1.5116}}  \\ \hline
		\end{tabular}
	}	\vspace{-1em}
	\caption{ Numerical comparisons of the enhancement on stripe noise  with seven advanced competitors. }	\vspace{-1em}~\label{tab:stripe}
\end{table*}

\begin{table*}[tb]
	\centering
	\footnotesize
	\renewcommand{\arraystretch}{1.1}
	
	\setlength{\tabcolsep}{2.8mm}{
		\begin{tabular}{|c|c|c|c|c|c|c|c|c|c|c|c|}
			\hline
			Scale                   & Metric & LINF & CRAFT & HAT    & ETDS   & BTC    & KXNet  & SwinIR & SR-LUT & FeMaSR & Ours   \\ \hline
			\multirow{5}{*}{$\times$2} & MI     &\cellcolor{blue!10} {2.9892}    & 2.9731   & 2.9683 & 2.9696 & 2.9794 & 2.9839 & 2.9723 & 2.9666 & 2.9397 &\cellcolor{red!15}{\textbf{3.0536}} \\ \cline{2-12}
			& VIF    & \cellcolor{red!15}{\textbf{1.0676}}    & 1.0651   & 1.0646 & 1.0646 &\cellcolor{blue!10} 1.0662 & 1.0627 & 1.0651 & 1.0629 & 1.0519 & 1.0511 \\ \cline{2-12}
			& SD     & 10.265    & 10.263   & 10.264 & 10.263 & 10.262 & 10.264 & 10.263 &\cellcolor{blue!10} {10.269} & 10.251 &\cellcolor{red!15}\textbf{10.495} \\ \cline{2-12}
			& $\mathrm{Q^{AB/F}}$   & \cellcolor{blue!10} {0.5133}    & 0.5041   & 0.5022 & 0.5026 & 0.5067 &\cellcolor{red!15} 0.5147 & 0.5028 & 0.5028 & 0.4985 &0.4698 \\ \cline{2-12}
			& EN     & 6.6534    & 6.6572   & 6.6579 & 6.6571 & 6.6544 & 6.6574 & 6.6572 & \cellcolor{blue!10}6.6631 & 6.6395 &\cellcolor{red!15}\textbf{6.8261} \\ \cline{1-12}
			\multirow{5}{*}{$\times$4}  & MI     & \cellcolor{blue!10}3.0031    & 2.9958   & 2.9966 & 2.9901 & 3.0020 & 2.9738 & 2.9941 & 2.9793 & 2.8927 &\cellcolor{red!15}\textbf{3.1162} \\ \cline{2-12}
			& VIF    &\cellcolor{blue!10} 1.0608    & 1.0482   & 1.0504  & 1.0449 & 1.0481 & 1.0549 & 1.0482 & 1.0449 & 1.0268 &\cellcolor{red!15}\textbf{1.0648} \\ \cline{2-12}
			& SD     & 10.267    & 10.264   & 10.259 & 10.266 & 10.251 & 10.268 & 10.263 & 10.263 &\cellcolor{blue!10} 10.306 &\cellcolor{red!15}\textbf{10.511} \\ \cline{2-12}
			& $\mathrm{Q^{AB/F}}$   & 0.529    & \cellcolor{red!15}\textbf{0.5303}  & 0.5275 & \cellcolor{blue!10} 0.5296 & 0.5284 & 0.5055 & 0.5295 & 0.5220 & 0.4774 &0.4996 \\ \cline{2-12}
			& EN     & 6.6659    & 6.6764   & 6.6772 & 6.6721 & 6.6730 & 6.6704 & 6.6767 & 6.6745 & \cellcolor{blue!10} 6.6903 &\cellcolor{red!15}{\textbf{6.8287}}  \\ \hline
		\end{tabular}
	}	\vspace{-1em}
	\caption{ Numerical comparisons of image super-resolution task with nine state-of-the art methods. }	\vspace{-1em}~\label{tab:super_resolution}
\end{table*}
\begin{figure*}[htb]
	\centering
	\setlength{\tabcolsep}{1pt}
	\begin{tabular}{c}
		
		\includegraphics[width=0.99\textwidth]{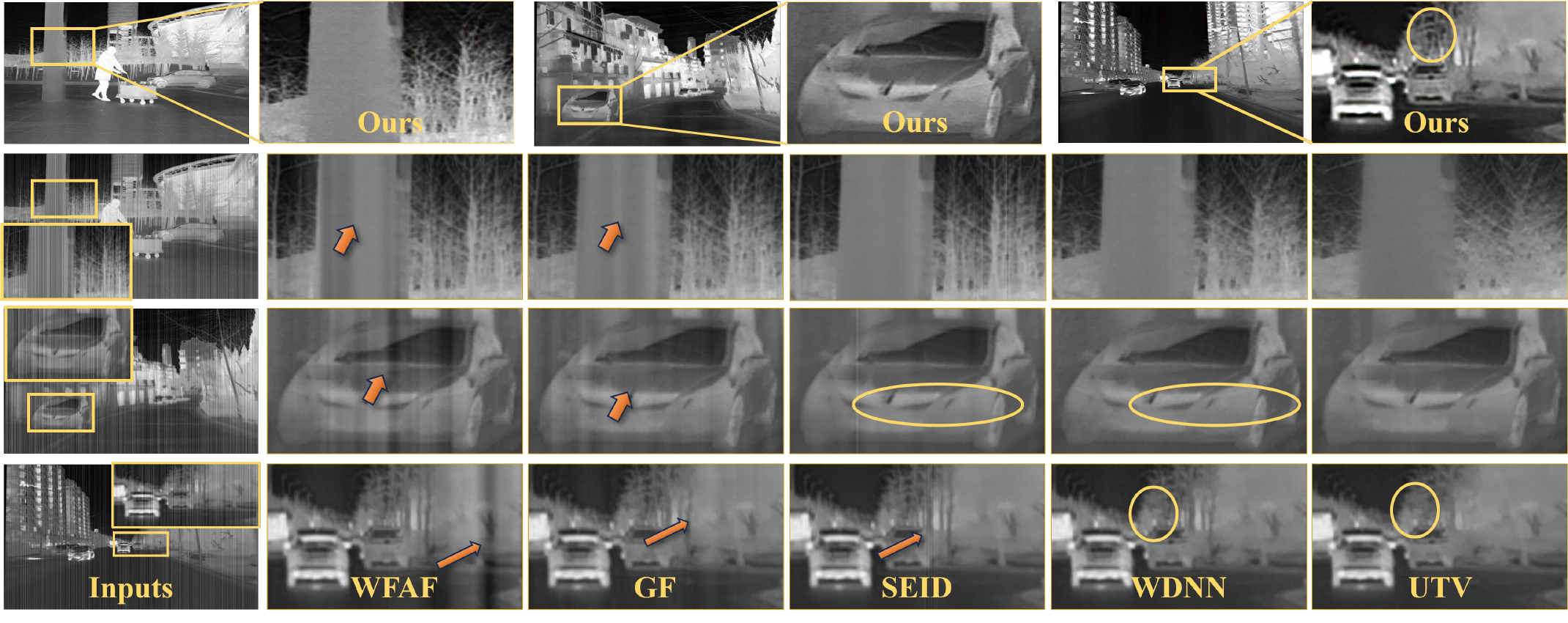}
		
	\end{tabular}
	\vspace{-1em}
	\caption{Visual comparisons of stripe noise removing with several advanced competitors.}
	\label{fig:noise}
\end{figure*}

\subsection{Dual Interaction Network}~\label{sec:din}
We propose a dual interaction network, composed of paired operations, including scale transform and spiking-guided separation.
The architecture is plotted in Fig.~\ref{fig:illustration} (b).

\noindent\textbf{Scale transform module.} Scale transform is widely leveraged for image restoration tasks~\cite{liu2024coconet}, which can effectively enhance the spatial structure and remove noise and artifacts.  We construct a Scale Transform Module (STM) to further fine contextual information at the feature level. We utilize the up-sampling and down-sampling  with one convolution operations to scale features with dense connections, following with literature~\cite{liu2019dual}.

\noindent\textbf{Spiking-guided separation module.}
We observe that the pixel intensity of degradation, such as infrared stripe noise, is typically higher than in other regions of the images. This type of noise usually results from inconsistencies or calibration errors in the response to infrared wavelengths, leading to anomalous brightness or intensity levels. In Spiking Neural Networks (SNNs), information is encoded as discrete spikes (e.g., binary signals), which can be leveraged to align the thermal-specific noise pattern more effectively by causing more neurons to fire. Thus, we propose a Spiking-Guided Separation Module (SSM) to isolate degraded features. As shown in Fig.~\ref{fig:illustration} (c), it comprises three components: Leaky Integrate-and-Fire (LIF) neurons to convert spatial features into spike sequences, standard convolution, and threshold-dependent batch normalization. These paired operations can ensure compact parameters, low energy consumption, and effective analysis of thermal degradation.

\section{Experiment Results}
In this section, we first elaborate on the concrete experimental settings. Next, we provide comprehensive comparisons for single corruption. Following that, we conduct experiments under general infrared imaging benchmarks. Lastly, we perform sufficient ablation studies on learning strategies and architectures. Details of the degradation generation are provided in the supplementary material.
% In this section, we first elaborate the concrete experimental settings.  Then, we make comprehensive comparisons for single corruption. Then we conduct the experiments under general infrared imaging benchmarks.  Lastly, sufficient ablation studies of  learning strategies and architectures are performed. Details of the degradation generation are provided in the supplementary material.

%		In this part,  we make a comparison with various state-of-the-art methods on the diverse  corruption factors. Then, sufficient ablation studies for  learning strategies and networks are performed.

% We used the PyTorch framework to implement  models on a single A40 GPU. Degradation model was optimized using  SGD, while enhancement model utilized  Adam. The learning rates $\gamma_\mathtt{e}$ and $\gamma_\mathtt{g}$ were  set to $1e^{-4}$ and $2e^{-4}$. We trained on 3600 images from M3FD~\cite{liu2022target} for approximately 50 epochs with warm start of 5 epochs. We actually consider the original images as $\mathbf{x}$ and $\mathbf{y}$, respectively.
We used the PyTorch framework to implement models on a single V100 GPU. The degradation model was optimized using  SGD, while the enhancement model utilized  Adam Optimizer. The learning rates $\gamma_\mathtt{e}$ and $\gamma_\mathtt{g}$ were  set to $1e^{-4}$ and $2e^{-4}$. We only trained on 50 images from M3FD~\cite{liu2022target} for 840 epochs. 
The total loss function $\mathcal{L}$ in Eq.~\eqref{eq:main2} and Eq.~\eqref{eq:constraint2}, composited by the measurement of pixel intensities $\mathcal{L}_\mathtt{pixel}$ and structural similarity $\mathcal{L}_\mathtt{SSIM}$. 
The whole formulation of the criterion can be written as:
$\mathcal{L} (\hat{\mathbf{y}};\mathbf{y}) = \alpha \mathcal{L}_\mathtt{pixel}(\hat{\mathbf{y}};\mathbf{y})+ \beta \mathcal{L}_\mathtt{SSIM}(\hat{\mathbf{y}};\mathbf{y}),
$
where 
$\alpha$ and $\beta$ are the trade-off parameters of the related loss functions. In this paper, we set $\alpha$ and $\beta$ as 0.75 and 1.1, respectively. As for the degradation generation, we leverage the  negative losses $-\mathcal{L} (\hat{\mathbf{y}};\mathbf{y})$ with  $\mathcal{L} (\hat{\mathbf{x}};\mathbf{x})$ for optimization.

\begin{figure*}[thb]
	\centering \begin{tabular}{c@{\extracolsep{0.25em}}c}
		\includegraphics[width=0.48\textwidth]{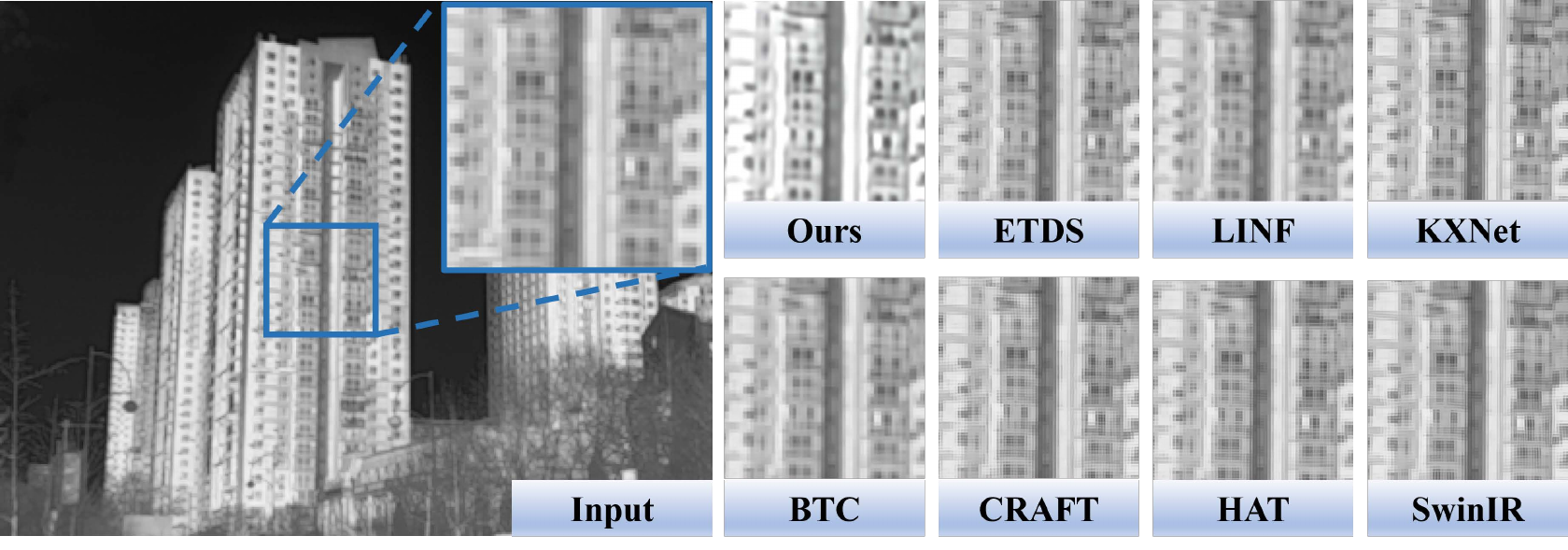}
		&	\includegraphics[width=0.48\textwidth]{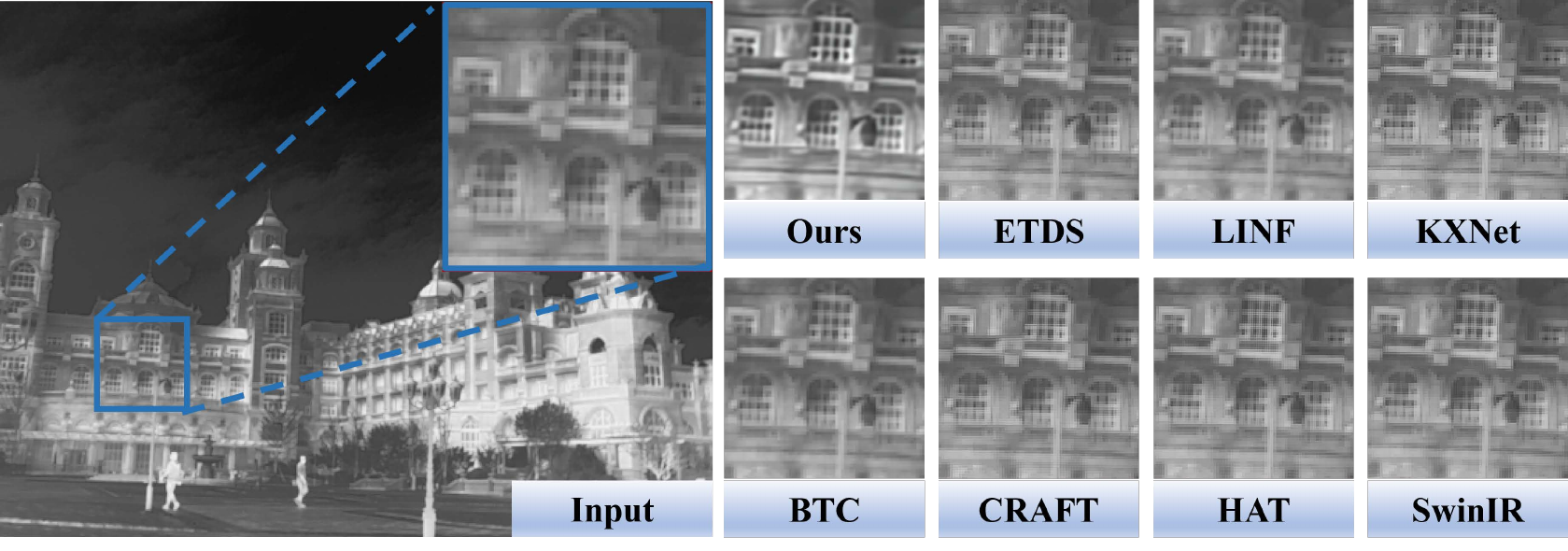}\\
		
		\includegraphics[width=0.48\textwidth]{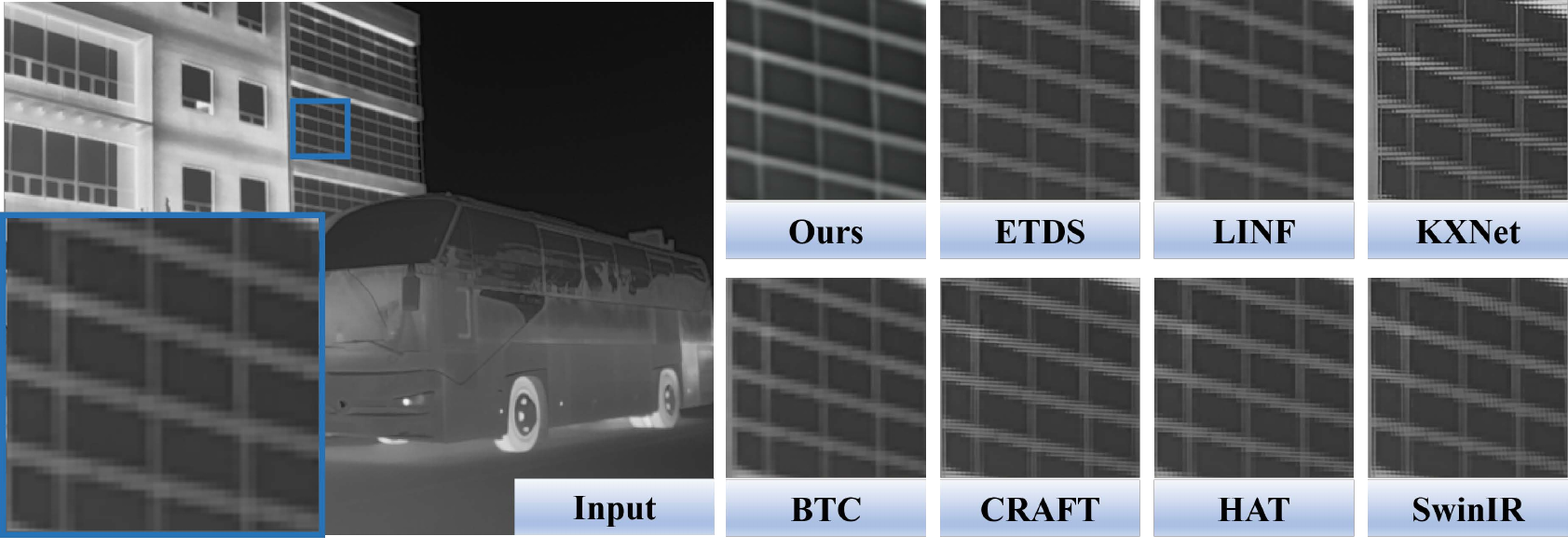}
		&	\includegraphics[width=0.48\textwidth]{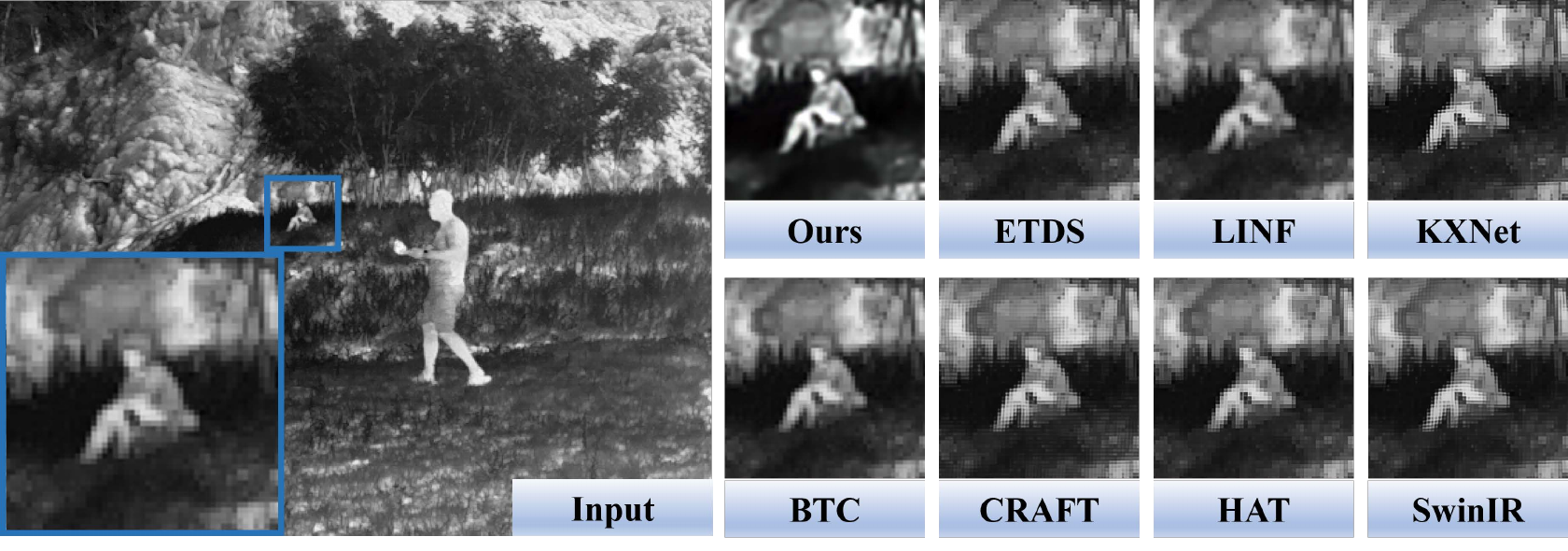}\\
		
	\end{tabular}
	\vspace{-1em}
	\caption{Infrared image super-resolution compared with several state-of-the-art  methods.}
	\label{fig:result_sr}
\end{figure*}

\subsection{Comparisons on Single Degradation}
Note that our method simultaneously addresses diverse degradation factors. It is challenging to establish appropriate ground truths and utilize widely-used reference-based metrics (e.g., PSNR and SSIM) to objectively measure performance with specialized schemes fairly. Thus we introduce the measurements of fusion for numerical evaluation.  

%MI  measures the amount of information shared between distinct modalities, indicating the information preservation ability. VIF  quantifies the visual information retained in fused images  based on the theory of natural scene statistics. SCD measures the pixel-wise correlation with source images, illustrating the influences to maintain the overall structure and features. SD quantifies the variation or dispersion of pixel values in the fused image, providing insights into the overall intensity distribution and contrast enhancement. $\mathrm{Q^{AB/F}}$ assesses the similarity between edges extracted from fused images and source inputs, particularly emphasizing aspects like local contrast, gradient distribution, and spectral information.

\textbf{Stripe noise restoration.} In this part, we introduce seven competitive methods for comparison, including MaskedDN~\cite{chen2023masked}, GF~\cite{cao2015effective}, SEID~\cite{song2023simultaneous}, WDNN~\cite{barral2024fixed}, WFAF~\cite{munch2009stripe}, UTV~\cite{bouali2011toward}, and LRSID~\cite{chang2016remote} on the stripe noise scenes.  
The subjective results under two degrees of stripe noise are reported in Table~\ref{tab:stripe}. 
We adopt  ReCo~\cite{huang2022reconet} as a basic fusion method for a fair comparison. 
Three remarkable advantages of our proposed method can be concluded from this table. Firstly, our scheme can effectively restore the inherent characteristics of infrared images with high data fidelity with the highest MI. Secondly, our method generates visual-appealing results (higher VIF and $\mathrm{Q^{AB/F}}$), which is consistent with the human visual perception system. Lastly, the results of our method illustrate the high contrast effects, indicated by the highest SD and SCD.
Moreover, visual comparisons are depicted in Fig.\ref{fig:noise}, showcasing three challenging scenes:  dense textures, low constrast, and weak  details.  It is noticeable that GF and WDNN cannot effectively preserve the sharp structure and smooth textural details.
Our method presents two key advantages: it effectively removes stripe noise without leaving visible residues, as seen in smooth  surfaces, clearly enhances thermal radiation information with well-preserved structural details, such as the shapes of pedestrians and forest.
%We also provide the edge maps based on Sobel operation to better showcase the noise removal and restoration of vertical details.

%From these comparisons, three notable observations emerge. Firstly, our method effectively removes stripe noise without significant residues, as evidenced by the smooth road surface in the first row. In contrast, methods such as GF, MaskedDN, and SEID struggle to eliminate strip streaks visibly. The obvious artifacts are still contained in these scenarios.
%Secondly, our approach remarkably highlights thermal radiation information with a clear structure, exemplified by the pedestrians shown in the second row. It is noticeable that GF and WDNN cannot effectively preserve the sharp structure and smooth textural details, such as the shapes of cars and pedestrians. 
%Lastly, our method can simultaneously reduce the noise effects and maintain the vertical texture (\textit{i.e.,} forest) effectively.

\begin{table*}[tb]
	\centering
	\footnotesize
	\renewcommand{\arraystretch}{1.1}
	
	\setlength{\tabcolsep}{1mm}{
		\begin{tabular}{|c|ccccc|ccccc|}
			\hline
			Datasets	& \multicolumn{5}{c|}{TNO}                                                                                                                                & \multicolumn{5}{c|}{RoadScene}                                                                                                                          \\ \hline
			Metrics             & \multicolumn{1}{c|}{LINF+SEID} & \multicolumn{1}{c|}{LINF+UTV} & \multicolumn{1}{c|}{KXNet+SEID} & \multicolumn{1}{c|}{KXNet+UTV} & Ours            & \multicolumn{1}{c|}{LINF+SEID} & \multicolumn{1}{c|}{LINF+UTV} & \multicolumn{1}{c|}{KXNet+SEID} & \multicolumn{1}{c|}{KXNet+UTV} & Ours            \\ \hline
			MI                  & \multicolumn{1}{c|}{\cellcolor{blue!10}{2.2107}}      & \multicolumn{1}{c|}{2.1300}     & \multicolumn{1}{c|}{2.1992}     & \multicolumn{1}{c|}{2.0724}    &\cellcolor{red!15}{\textbf{2.3950}} & \multicolumn{1}{c|}{\cellcolor{blue!10}{3.2690}}      & \multicolumn{1}{c|}{3.1268}     & \multicolumn{1}{c|}{3.2625}     & \multicolumn{1}{c|}{3.1232}    & \cellcolor{red!15}{\textbf{3.3477}} \\ \hline
			SD                  & \multicolumn{1}{c|}{\cellcolor{blue!10}{8.9578}}    & \multicolumn{1}{c|}{8.8680}     & \multicolumn{1}{c|}{8.8822}     & \multicolumn{1}{c|}{8.7715}    &\cellcolor{red!15}{\textbf{9.1313}} & \multicolumn{1}{c|}{\cellcolor{blue!10}{10.321}}      & \multicolumn{1}{c|}{10.201}     & \multicolumn{1}{c|}{10.300}     & \multicolumn{1}{c|}{10.179}    &\cellcolor{red!15}{\textbf{10.660}} \\ \hline
			$\mathrm{Q^{AB/F}}$ & \multicolumn{1}{c|}{\cellcolor{blue!10}{0.3888}}      & \multicolumn{1}{c|}{0.3562}     & \multicolumn{1}{c|}{0.3854}     & \multicolumn{1}{c|}{0.3410}    & \cellcolor{red!15}{\textbf{0.4176}} & \multicolumn{1}{c|}{0.4243}      & \multicolumn{1}{c|}{0.4187}     & \multicolumn{1}{c|}{\cellcolor{blue!10}{0.4246}}       & \multicolumn{1}{c|}{0.4194}    & \cellcolor{red!15}{\textbf{0.4470}} \\ \hline
			SCD                 & \multicolumn{1}{c|}{1.3480}      & \multicolumn{1}{c|}{1.3635}     & \multicolumn{1}{c|}{1.3576}     & \multicolumn{1}{c|}{\cellcolor{blue!10}{1.3643}}    & \cellcolor{red!15}{\textbf{1.4454}} & \multicolumn{1}{c|}{1.5411}      & \multicolumn{1}{c|}{\cellcolor{blue!10}{1.5938}}     & \multicolumn{1}{c|}{1.5433}     & \multicolumn{1}{c|}{1.5935}    & \cellcolor{red!15}{\textbf{1.6131}} \\ \hline
			EN                  & \multicolumn{1}{c|}{\cellcolor{blue!10}{6.6543}}      & \multicolumn{1}{c|}{6.6116}     & \multicolumn{1}{c|}{6.6393}     & \multicolumn{1}{c|}{6.6135}    & \cellcolor{red!15}{\textbf{6.7835}} & \multicolumn{1}{c|}{6.9861}      & \multicolumn{1}{c|}{6.9466}     & \multicolumn{1}{c|}{\cellcolor{blue!10}{6.9884}}     & \multicolumn{1}{c|}{6.9473}    & \cellcolor{red!15}{\textbf{7.1000}} \\ \hline
		\end{tabular}
	}	\vspace{-1em}
	\caption{ Numerical comparisons of joint enhancement under two general benchmarks. }	\vspace{-1em}~\label{tab:general}
\end{table*}
\begin{table}[tb]
	\centering
	\footnotesize
	\renewcommand{\arraystretch}{1.1}
	
	\setlength{\tabcolsep}{1.4mm}{
		\begin{tabular}{|c|c|c|c|c|c|c|}
			\hline
			Fusion                & Metrics & AirNet & DALE   & IAT    & MBLLEN & Ours   \\ \hline
			\multirow{5}{*}{ReCo} & MI     & 2.4599 & 1.7566 &\cellcolor{blue!10} 2.8294 & 2.7484 &\cellcolor{red!15}\textbf{3.1636} \\  \cline{2-7}
			& VIF    &\cellcolor{blue!10}  1.0319 & 0.9762 & 0.9980 &\cellcolor{red!15} \textbf{1.0486} & \ 0.9046 \\  \cline{2-7}
			& SD     & 9.0136 & 8.5928 & 9.3036 &\cellcolor{blue!10}\ 9.7832 &\cellcolor{red!15} \textbf{10.7724} \\  \cline{2-7}
			& $\mathrm{Q^{AB/F}}$   & 0.4772 & 0.4850 &\cellcolor{blue!10} 0.5030 &\cellcolor{red!15} \textbf{0.5357} &\ 0.5001 \\  \cline{2-7}
			& EN     & 6.0448 & 6.0811 & \cellcolor{blue!10}6.6712 & 6.5364 &\cellcolor{red!15}\textbf{6.7142} \\  \hline
			\multirow{5}{*}{LRR~\cite{li2023lrrnet}}  & MI     & \cellcolor{red!15} \textbf{3.1392} & 2.9433 &\cellcolor{blue!10} 3.0799 & 2.9304 &3.0118 \\  \cline{2-7}
			& VIF    & 0.7390 & 0.8086 &\cellcolor{blue!10} 0.8155 & 0.7843 &\cellcolor{red!15}\textbf{0.8200} \\  \cline{2-7}
			& SD     & 7.6213 & 7.9167 &\cellcolor{blue!10} 8.0923 & 7.7350 &\cellcolor{red!15} \textbf{9.9614} \\  \cline{2-7}
			& $\mathrm{Q^{AB/F}}$   & 0.4680 & 0.4463 & 0.4915 & \cellcolor{red!15}\textbf{0.5330}&{\cellcolor{blue!10}0.5219} \\  \cline{2-7}
			& EN     & 6.3720 & 6.1191 &\cellcolor{blue!10}{6.4803} & 6.3241 & \cellcolor{red!15}\textbf{6.5596} \\ \hline
		\end{tabular}
	}	\vspace{-1em}
	\caption{ Numerical results of low contrast enhancement tasks. }	\vspace{-1em}~\label{tab:constrast}
\end{table}
\begin{figure}[htb]
	\centering
	\setlength{\tabcolsep}{1pt}
	\begin{tabular}{ccc}
		\includegraphics[width=0.15\textwidth]{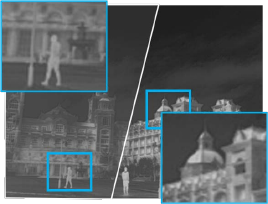}
		&\includegraphics[width=0.15\textwidth]{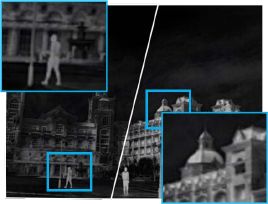}
		&\includegraphics[width=0.15\textwidth]{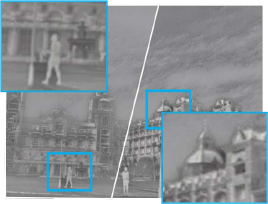}
		
		\\
		\footnotesize	 Original & \footnotesize AirNet & \footnotesize DALE
		\\
		\includegraphics[width=0.15\textwidth]{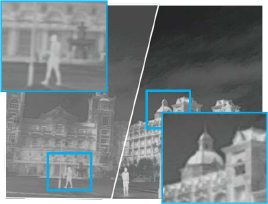}
		&\includegraphics[width=0.15\textwidth]{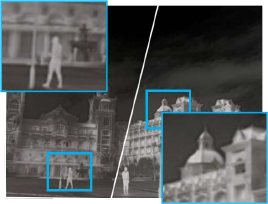}
		&\includegraphics[width=0.15\textwidth]{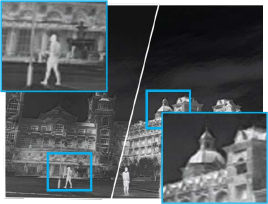}
		
		\\
		
		\footnotesize	 IAT & \footnotesize MBLLEN & \footnotesize Ours
		%				\\	
		%				\includegraphics[width=0.15\textwidth]{./pic/contrast/2/input}
		%				\includegraphics[width=0.15\textwidth]{./pic/contrast/2/AirNet}
		%				&\includegraphics[width=0.15\textwidth]{./pic/contrast/2/DALE}
		%				
		%				\\
		%				\footnotesize	 Original & \footnotesize AirNet & \footnotesize DALE
		%				\\
		%				\includegraphics[width=0.15\textwidth]{./pic/contrast/2/IAT}
		%				&\includegraphics[width=0.15\textwidth]{./pic/contrast/2/MBLLEN}
		%				&\includegraphics[width=0.15\textwidth]{./pic/contrast/2/ours}
		%		
		%				\\
		%		
		%				\footnotesize	 IAT & \footnotesize MBLLEN & \footnotesize Ours
		\\
	\end{tabular}
	\vspace{-1em}
	\caption{Visual comparisons of low-contrast enhancement with several advanced competitors under different corrupted degrees.}
	\label{fig:contrast}
\end{figure}

\textbf{Image super-resolution.} We compare nine latest state-of-the-art methods for image super-resolution, including LINF~\cite{yao2023local}, CRAFT~\cite{li2023feature}, HAT~\cite{chen2023activating}, ETDS~\cite{chao2023equivalent}, BTC~\cite{pak2023b}, KXNet~\cite{fu2022kxnet}, SwinIR~\cite{liang2021swinir}, SR-LUT~\cite{jo2021practical}, and FeMaSR~\cite{chen2022real}. The inputs are with same size.
Table~\ref{tab:super_resolution} presents numerical results  at $\times 2$, $\times 4$ resolution, respectively. In this study, we replaced the SCD with Entropy (EN) to enhance the measurement of textural details (modal information) present in the features. The table demonstrates the consistently superior performance of our method across five representative metrics. 
The informative metrics (MI and EN) underscore our method's ability to preserve more textural details in infrared images.  Moreover, the highest SD indicates a substantial enhancement in contrast across different fusion methods. 
The qualitative comparison under four challenging scenes is depicted in Fig.~\ref{fig:result_sr}. We can summarize two significant characteristics compared with existing methods. Firstly, most current schemes generate noticeable artifacts with blurred details caused by the bi-cubic downsampling. As shown in the third case, KXNet and CRAFT cannot effectively persevere the structures with lines.
Our method can effectively restore the high-frequency textural details (\textit{e.g.,} dense structure in the first case), which realizes the visual-appealing result and is beneficial for human observation. Furthermore, our method comprehensively 
considers the impacts of multi-degradation, which can significantly highlight the salient thermal objects with sharp structures  (\textit{e.g.,} street lamp in the second case). However, existing methods only consider one single degradation, which cannot effectively handle the contrast information.

\begin{table}[tb]
	\centering
	\footnotesize
	\renewcommand{\arraystretch}{1.1}
	
	\setlength{\tabcolsep}{1mm}{
		\begin{tabular}{|c|c|c|c|c|c|c|}
			\hline
			Data Usage & 20      & 50      & 100     & 150     & 200     & 350     \\ \hline
			MI         & 3.0373  & 3.0891  & 3.0393  & 3.1075  & 3.102   & 3.0891  \\ \hline
			VIF        & 0.8291  & 0.9310  & 0.8306  & 0.8343  & 0.8406  & 0.8441  \\ \hline
			SD         & 10.3819 & 10.3432 & 10.4543 & 10.3321 & 10.3129 & 10.4764 \\ \hline
			$\mathrm{Q^{AB/F}}$        & 0.445   & 0.4821  & 0.4454  & 0.4531  & 0.4555  & 0.4571  \\ \hline
			EN         & 6.811   & 6.8188  & 6.8014  & 6.8088  & 6.7923  & 6.7891  \\ \hline
		\end{tabular}
	}
	\vspace{-1em}\caption{Stripe denoising performance under diverse data usage. }	~\label{tab:data}
\end{table}	
\begin{figure}[thb]
	\centering
	\includegraphics[width=0.49\textwidth]{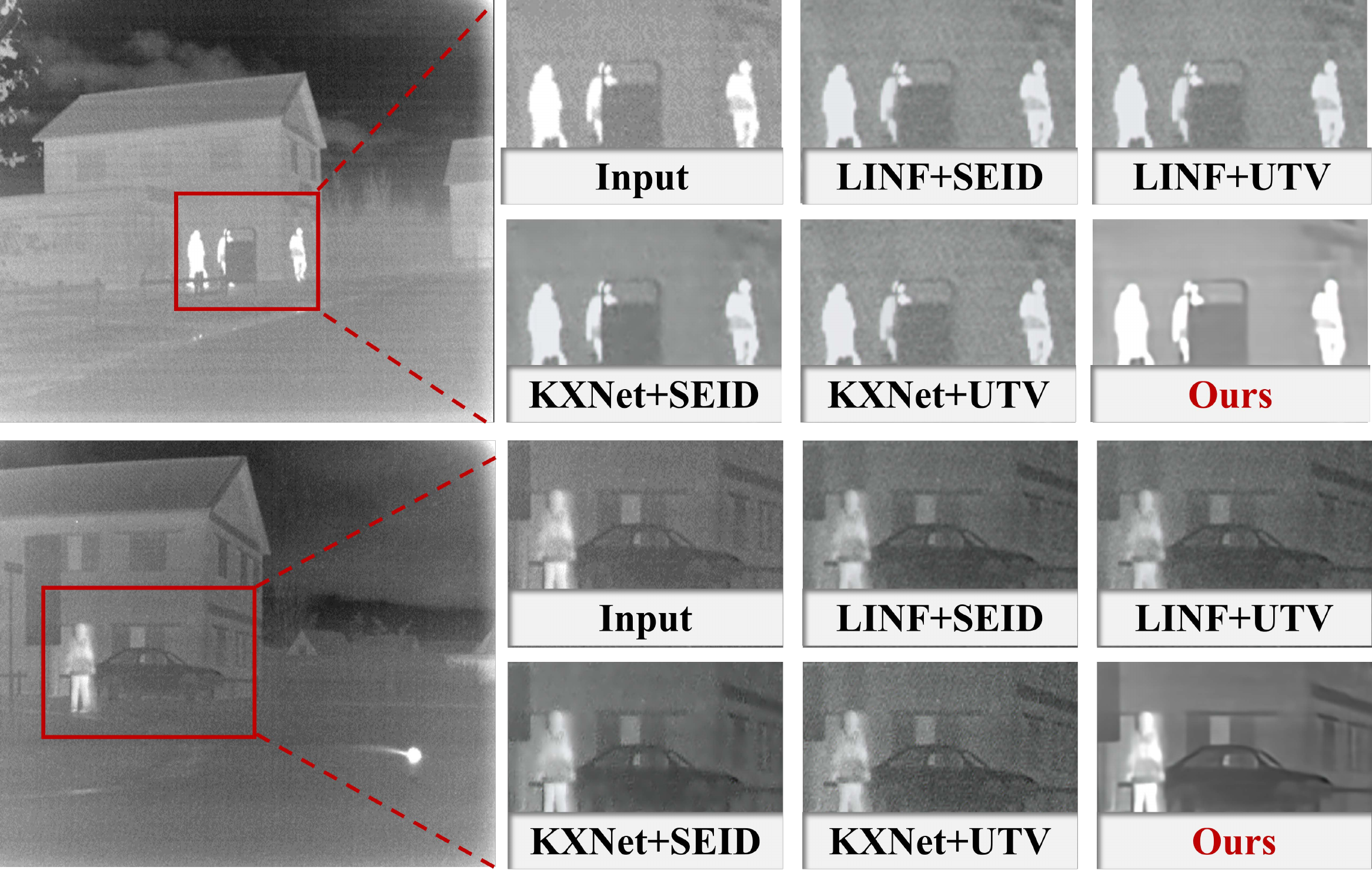}
	
	\caption{{Infrared enhancement on two challenging scenarios.
	}}
	
	\label{fig:general}
\end{figure}

\textbf{Low-contrast enhancement.} Moreover, we also compare our method under the low-contrast scenario. Due to there are a few methods designed for the low-contrast enhancement for infrared images, we adopt several all-in-one image restoration and contrast enhancement schemes for visible images, including AirNet~\cite{li2022all}, DALE~\cite{kwon2020dale}, IAT~\cite{cui2022you}, and MBLLEN~\cite{lv2018mbllen}.
Table~\ref{tab:constrast} presents the results of low-contrast enhancement compared with various methods. Remarkably, our method consistently demonstrates improvements across diverse metrics.
In summary, our approach effectively preserves edge information and textural details, resulting in high-contrast observations.
Furthermore, we provide two typical scenarios to illustrate the superiority in Fig.~\ref{fig:contrast}. Although the all-in-one method AirNet increases the contrast, it interferes with the infrared information. DALE and IAT improve the global intensity of the thermal image but introduce blurred texture details. In contrast, our scheme can effectively highlight thermal targets (e.g., pedestrians and building details) while preserving clear and sharp structures. This provides visually appealing observations that are consistent with human visual system.

%	\subsection{Comparisons on Combined Corruption}

\subsection{Comparisons on Composited Degradation}
We also demonstrate the effectiveness of our method  using  widely-used benchmarks TNO and RoadScene, which encompass diverse composited degradation factors. 
We select representative state-of-the-art methods for these corruptions to conduct a comparison, including UTV, SEID, LINF, and KXNet, uniformly enhanced by IAT and fused by ReCo. Table~\ref{tab:general} reports the numerical results. Specifically, the optimal MI and EN values indicate that our method effectively highlights thermal information in the fusion schemes. The highest SD value illustrates that our method  enhances contrast for human observation. Additionally, the highest $\mathrm{Q^{AB/F}}$ value demonstrates our method's remarkable ability to extract textural details and edge information.

Additionally, we present an illustrative visual comparison in Fig.~\ref{fig:general} featuring four challenging scenarios. These two examples are from the TNO dataset and are corrupted by heavy stripe noise and low resolution. Combined methods (\textit{e.g.,} LINF+SEID, KXNet+UTV) are unable to handle the complex noise, resulting in significant textural artifacts. Other combined schemes also exhibit residual artifacts due to the hybrid degradation. Clearly,  UTV-based schemes significantly degrade structure and textural details due to interference from low contrast factors. Importantly, our method effectively highlights thermal-sensitive objects while preserving sufficient textural details, representing a promising improvement in real-world thermal imaging.

%  Our method effectively removes real thermal
%noise as shown in the first row. Moreover, our method highlights
%thermal-salient objects with clear structure, as shown in the second
%row, demonstrating high applicability in perception systems.

\begin{table}[tb]
	\centering
	\footnotesize
	\renewcommand{\arraystretch}{1.1}
	
	\setlength{\tabcolsep}{1.3mm}{
		\begin{tabular}{|c|cc|ccc|}
			\hline
			\multirow{2}{*}{Operations} & \multicolumn{2}{c|}{Visual Metrics}  & \multicolumn{3}{c|}{Hardware Efficiency}                                     \\ \cline{2-6} 
			& \multicolumn{1}{c|}{VIF}    & Qabf   & \multicolumn{1}{c|}{Runtime(ms)} & \multicolumn{1}{c|}{FLOPs(G)} & Params.(M) \\ \hline
			Cascade-1                   & \multicolumn{1}{c|}{0.9929} & 0.5445 & \multicolumn{1}{c|}{80.84}   & \multicolumn{1}{c|}{303.4}    & 0.382      \\ \hline
			Cascade-2                   & \multicolumn{1}{c|}{0.9945} & 0.5545 & \multicolumn{1}{c|}{121.6}   & \multicolumn{1}{c|}{455.6}    & 0.572      \\ \hline
			Cascade-3                   & \multicolumn{1}{c|}{1.0001} & 0.5478 & \multicolumn{1}{c|}{331.7}   & \multicolumn{1}{c|}{762.4}    & 0.960      \\ \hline
		\end{tabular}
	}
	\vspace{-1em}\caption{Effectiveness of the paired operation. }	~\label{tab:effctiveness}
\end{table}	

\subsection{Ablation Study}
\begin{figure}[thb]
	\centering
	\includegraphics[width=0.49\textwidth]{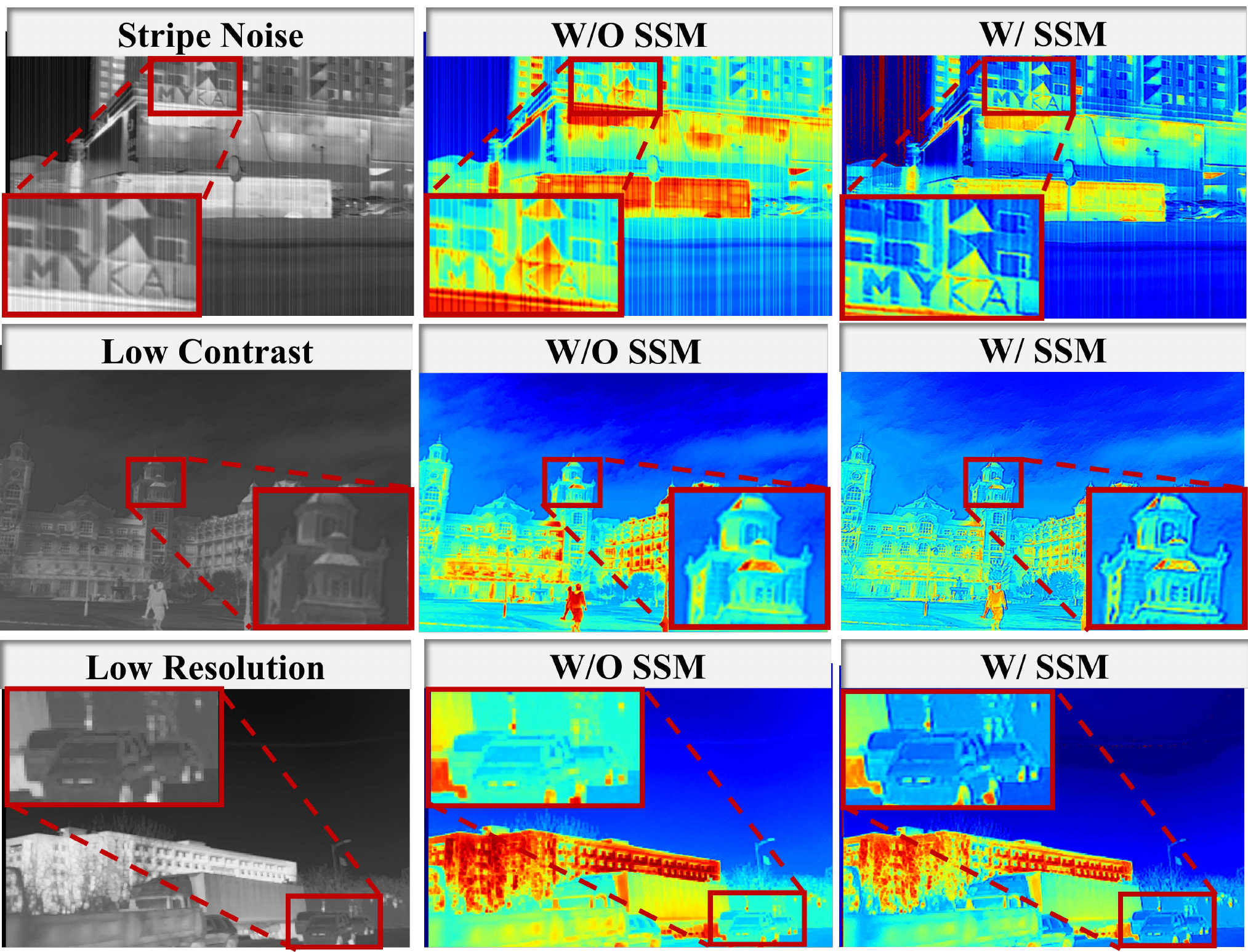}
	
	\caption{{ Feature visualization based on w/ and w/o SSM.}}
	
	\label{fig:feature}
\end{figure}
\begin{table}[tb]
	\centering
	\footnotesize
	\renewcommand{\arraystretch}{1.1}
	
	\setlength{\tabcolsep}{3mm}{
		\begin{tabular}{|c|c|c|c|c|}
			\hline
			Degradation                   & Metrics & Average & All & Proposed\\ \hline
			\multirow{2}{*}{Stripe Noise}   & VIF    & 0.7162   &\cellcolor{blue!10} 0.7467        &\cellcolor{red!15}\textbf{0.9310}          \\ \cline{2-5} 
			&  $\mathrm{Q^{AB/F}}$   & \cellcolor{blue!10}0.3965     & 0.3887        &\cellcolor{red!15}\textbf{0.4821}          \\ \hline
			\multirow{2}{*}{Low Resolution}       & VIF    & \cellcolor{blue!10}0.6698     & 0.6647        &\cellcolor{red!15}\textbf{0.8401}          \\ \cline{2-5} 
			&  $\mathrm{Q^{AB/F}}$   & \cellcolor{blue!10}0.4105     & 0.3929        &\cellcolor{red!15}\textbf{0.4458}          \\ \hline
			\multirow{2}{*}{Low Contrast}       & VIF    & 0.6530     & \cellcolor{blue!10} 0.6696        &\cellcolor{red!15}\textbf{0.8941}          \\ \cline{2-5} 
			&  $\mathrm{Q^{AB/F}}$   & \cellcolor{blue!10}0.4029     & 0.3880        &\cellcolor{red!15}\textbf{0.4924}          \\ \hline
			\multirow{2}{*}{Composited} & VIF    &\cellcolor{blue!10} 0.7171     & 0.7636        &\cellcolor{red!15}\textbf{0.9930}          \\ \cline{2-5} 
			&  $\mathrm{Q^{AB/F}}$   & 0.3967     & \cellcolor{blue!10}0.3971        &\cellcolor{red!15}\textbf{0.5445}          \\ \hline
		\end{tabular}
	}
	\vspace{-1em}\caption{ Numerical verification for  training strategies. }	~\label{tab:alb1}
\end{table}	

\noindent\textbf{Validation about the amounts of  training data.} Our method achieves excellent performance even with a limited amount of training data in Table~\ref{tab:data}. Specifically, with smaller datasets (\textit{e.g.}, 20 and 50 images),  metrics such as  VIF, and $\mathrm{Q^{AB/F}}$  already show relatively high values, indicating that our approach effectively enhances  quality even with limited data. 

\noindent\textbf{Verification of paired operations.} Table~\ref{tab:effctiveness} demonstrates the effectiveness of  paired operations in terms of visual metrics and hardware efficiency. Our method effectively balances model performance and computational efficiency by using a small number of cascades (only two paired operations in total), compared to other methods like BTC with 22.4M  and KXNet with 6.51M parameters, respectively.

\noindent\textbf{Effectiveness of training strategy.} We compared the proposed learning strategy with two alternative strategies: averaging the mix of degradations and training with all degradations  in Table~\ref{tab:alb1}.
Clearly, the averaged aggregation significantly improve performance compared to the original joint training with random selection. Moreover, employing dynamic degradation generation can further enhances performance under diverse corruptions.

\begin{figure}[htb]
	\centering
	\setlength{\tabcolsep}{1pt}
	\begin{tabular}{ccccc}
		
		\includegraphics[width=0.09\textwidth]{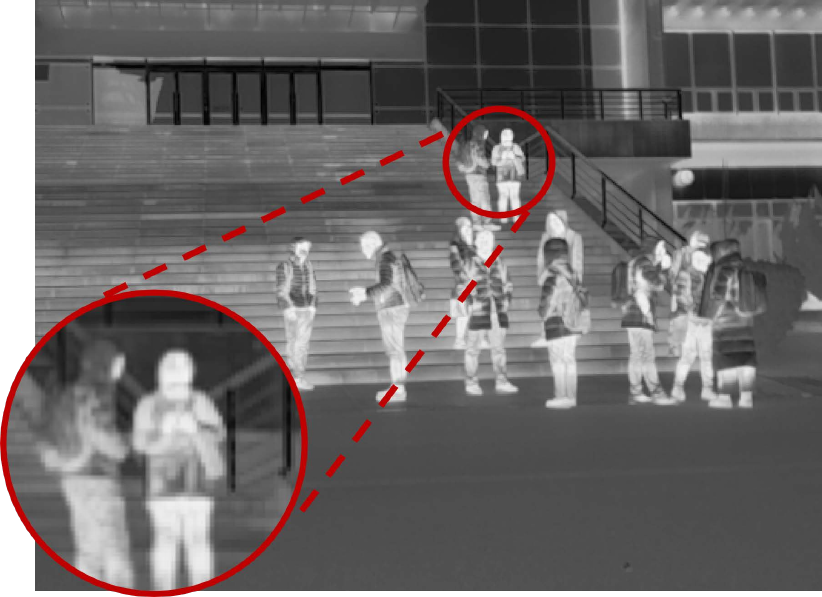}
		&\includegraphics[width=0.09\textwidth]{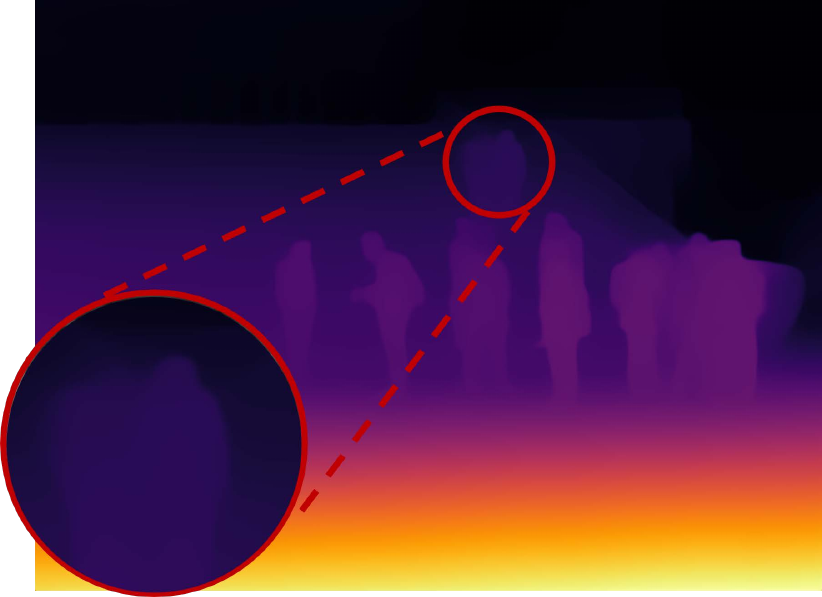}
		&\includegraphics[width=0.09\textwidth]{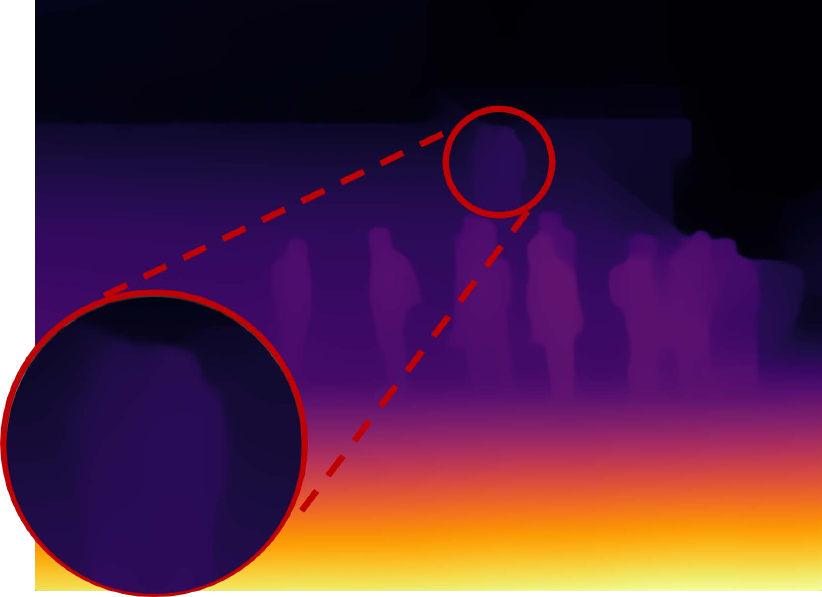}
		&\includegraphics[width=0.09\textwidth]{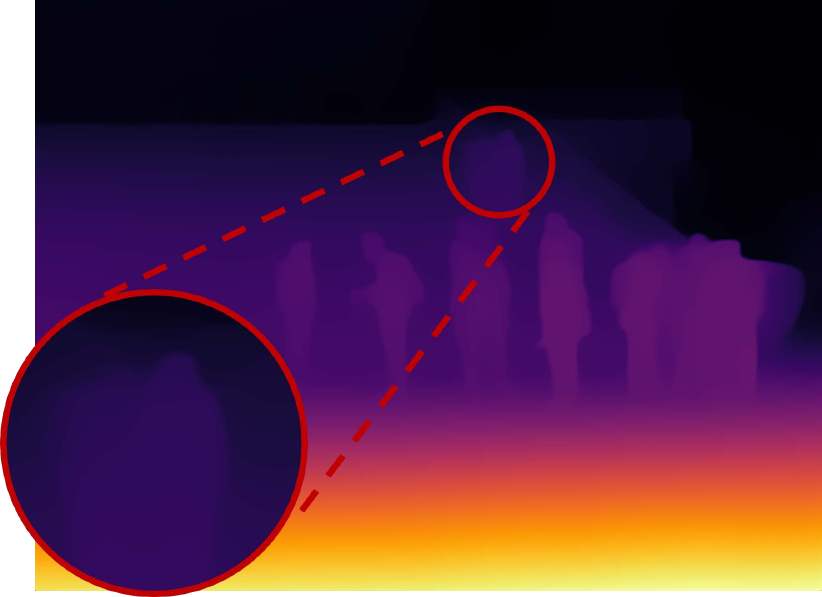}
		&\includegraphics[width=0.09\textwidth]{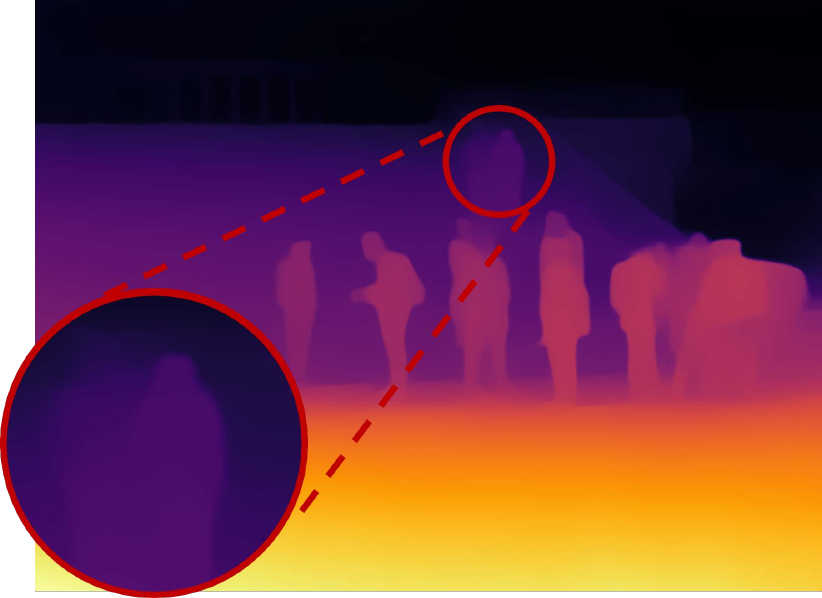}
		\\
		\includegraphics[width=0.09\textwidth]{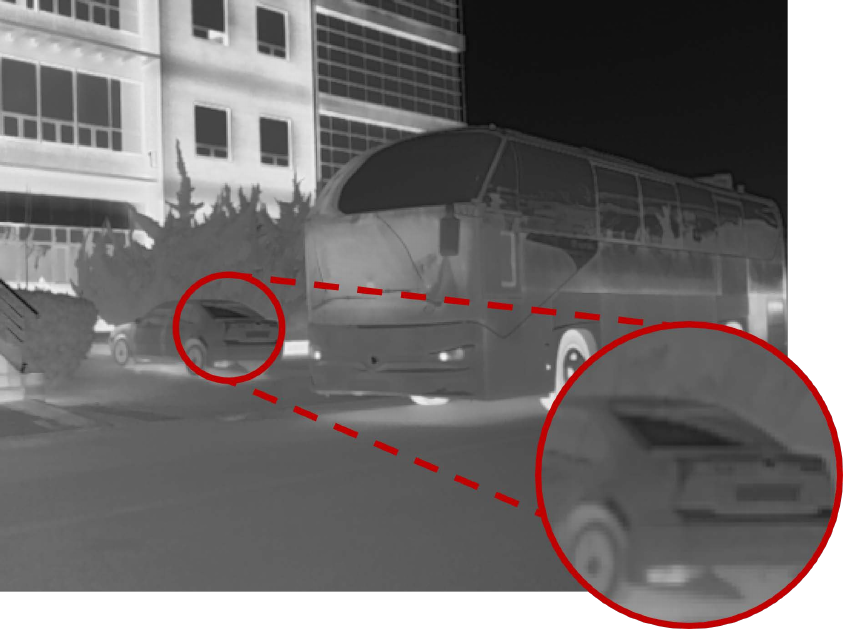}
		&\includegraphics[width=0.09\textwidth]{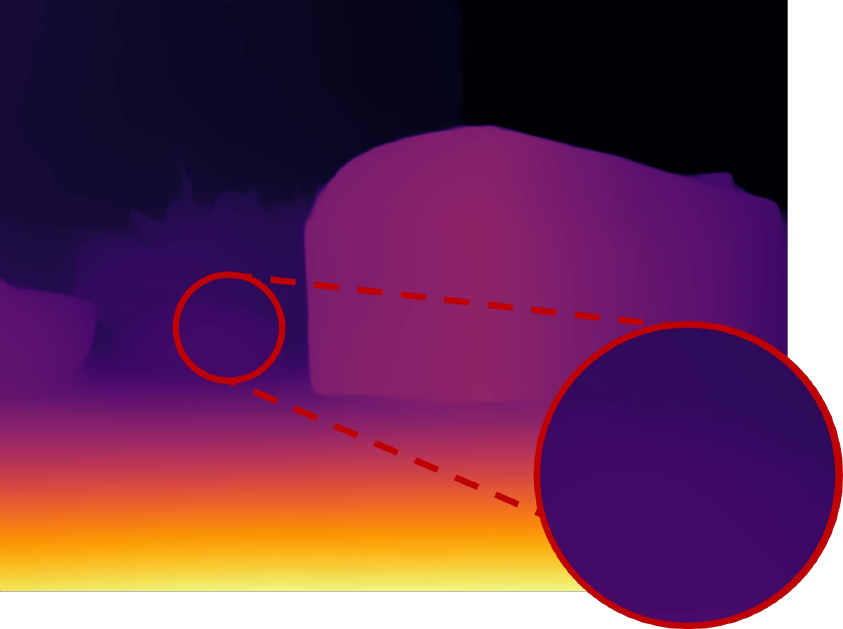}
		&\includegraphics[width=0.09\textwidth]{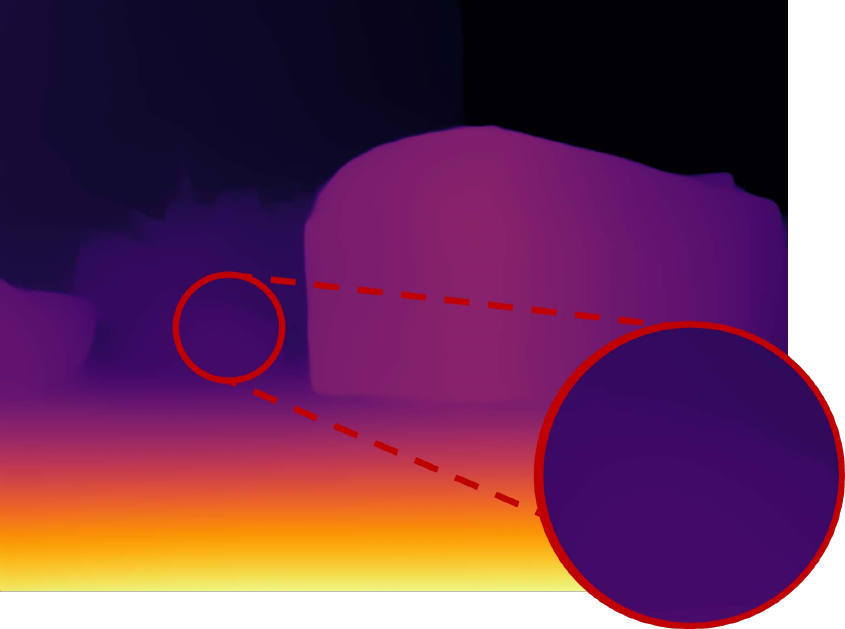}
		&\includegraphics[width=0.09\textwidth]{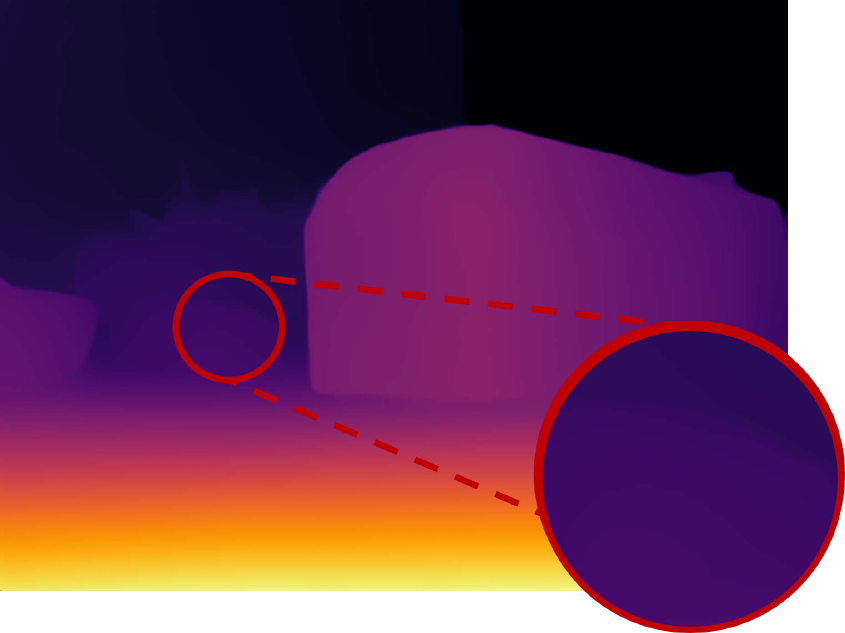}
		&\includegraphics[width=0.09\textwidth]{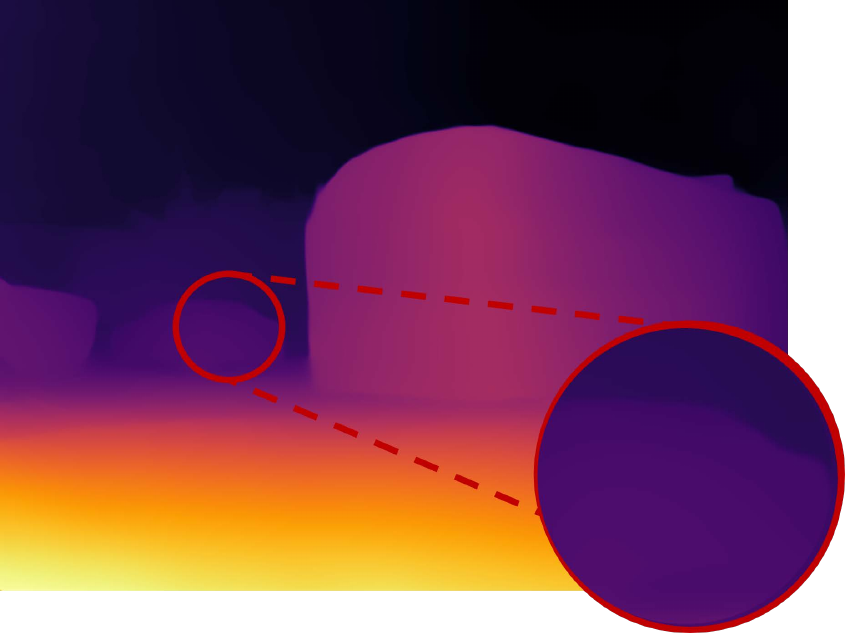}
		\\
		\includegraphics[width=0.09\textwidth]{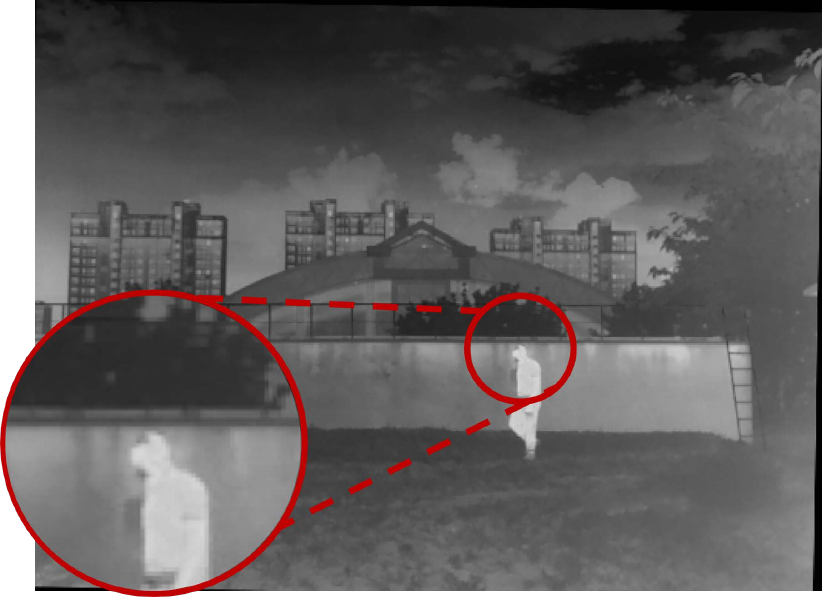}
		&\includegraphics[width=0.09\textwidth]{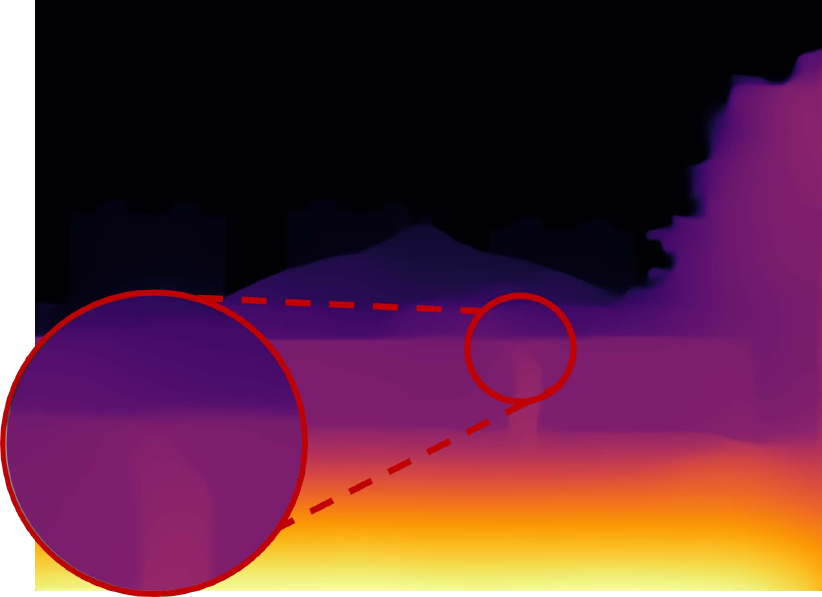}
		&\includegraphics[width=0.09\textwidth]{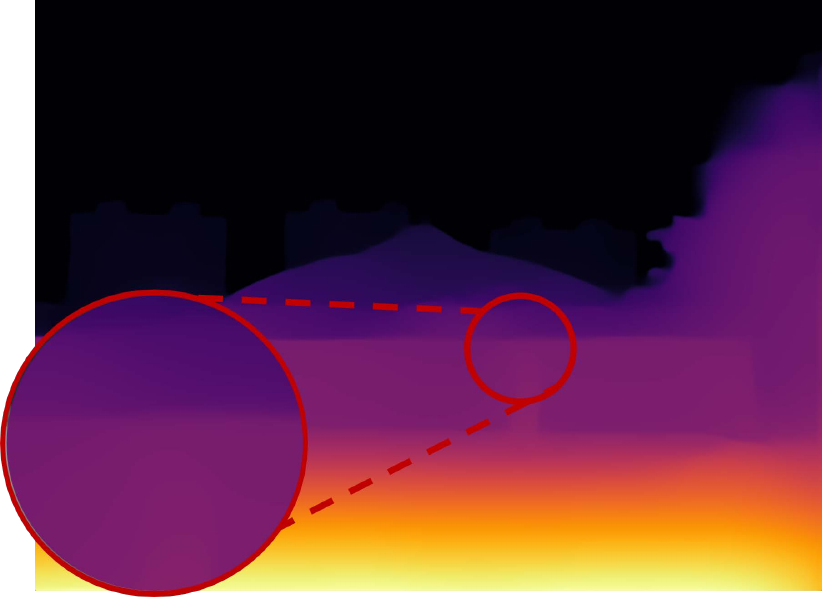}
		&\includegraphics[width=0.09\textwidth]{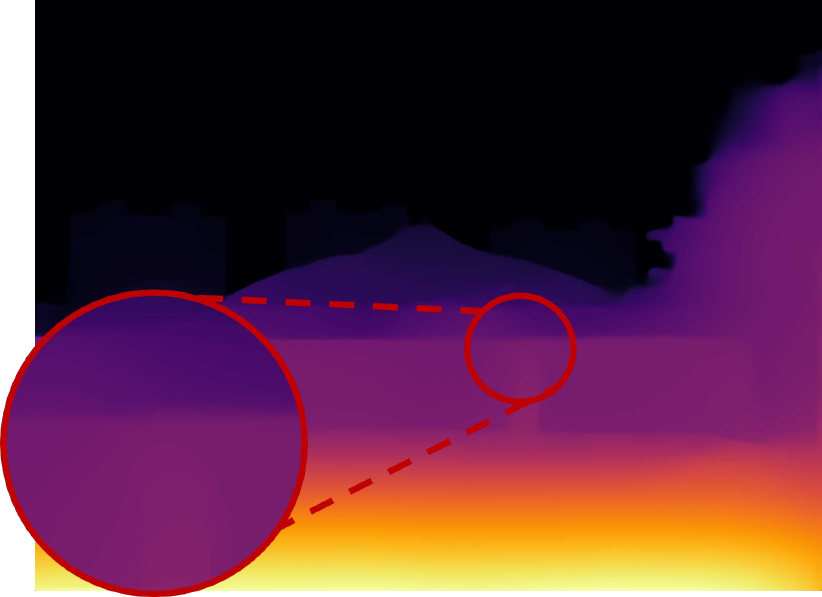}
		&\includegraphics[width=0.09\textwidth]{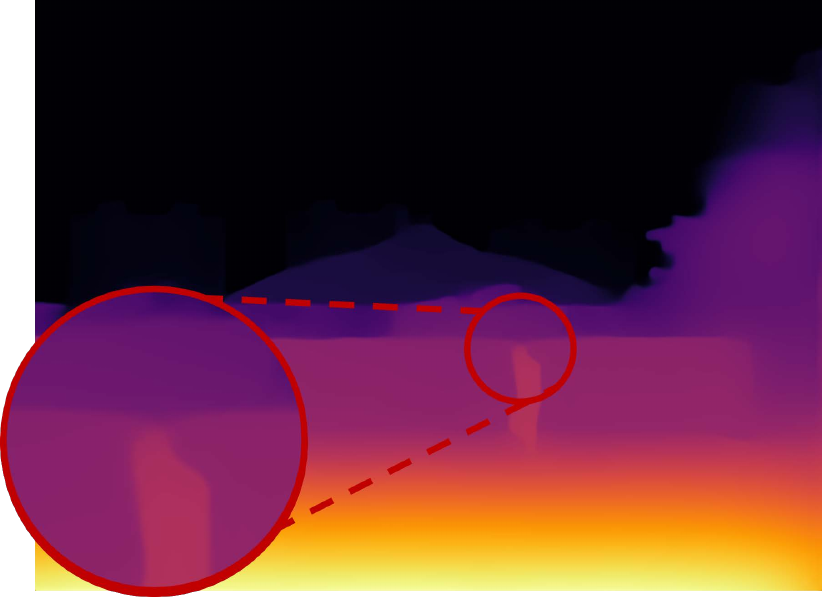}
		\\
		\footnotesize	 Input & \footnotesize BTC & \footnotesize CRAFT &\footnotesize ETDS &  \footnotesize Ours
		\\
	\end{tabular}
	\vspace{-1em}
	\caption{Depth estimation under low-resolution infrared images.}
	\label{fig:depth}
\end{figure}
\noindent\textbf{Validation of proposed architecture.}  We visualize the features with and without SSM in Fig.~\ref{fig:feature}, demonstrating a noticeable improvement in identifying specific degradation factors. For instance, utilizing the SSM can effectively identify the stripe streaks. Furthermore, the structure of bumps and buildings also can be remarkably highlighted for the restoration. These instances demonstrate our superiority.

\begin{figure}[htb]
	\centering
	\setlength{\tabcolsep}{1pt}
	\begin{tabular}{ccc}
		\includegraphics[width=0.15\textwidth]{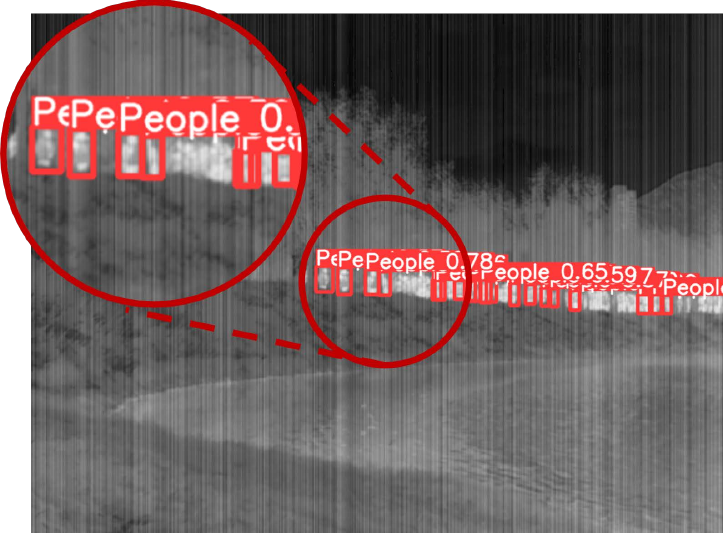}
		&\includegraphics[width=0.15\textwidth]{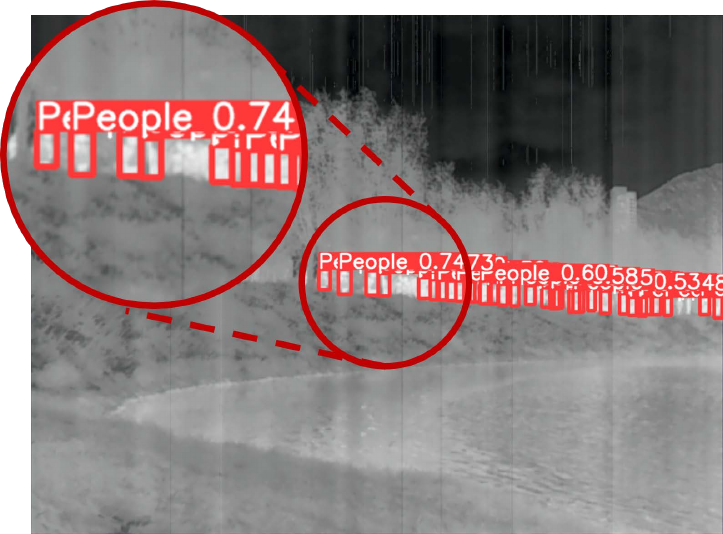}
		&\includegraphics[width=0.15\textwidth]{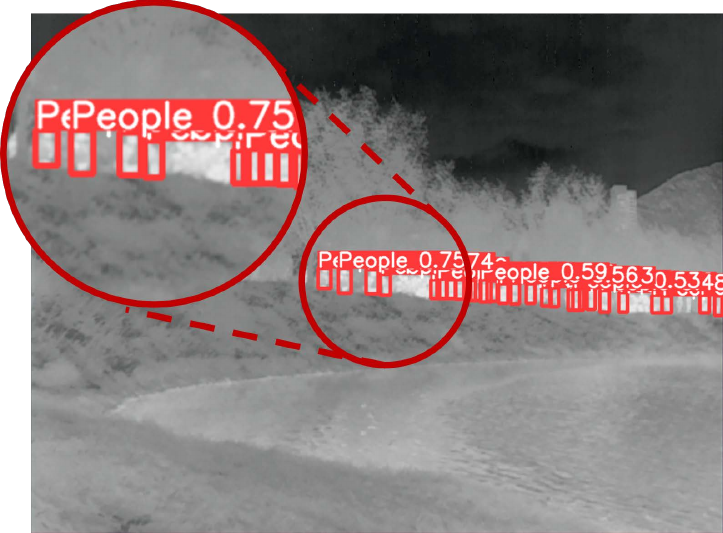}
		
		\\
		\footnotesize	 Degraded & \footnotesize SEID+LINF+IAT & \footnotesize SEID+KXNet+IAT
		\\
		\includegraphics[width=0.15\textwidth]{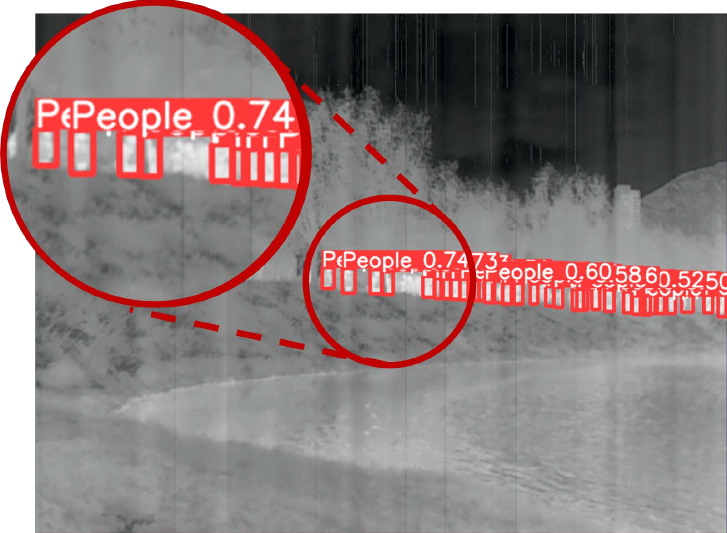}
		&\includegraphics[width=0.15\textwidth]{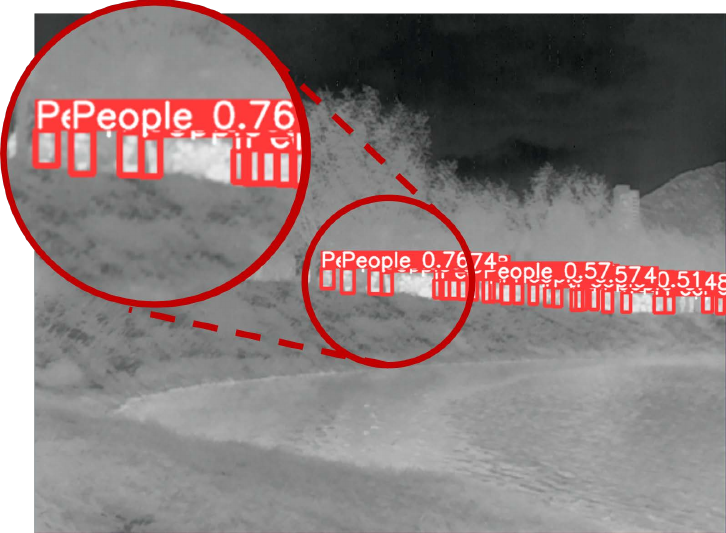}
		&\includegraphics[width=0.15\textwidth]{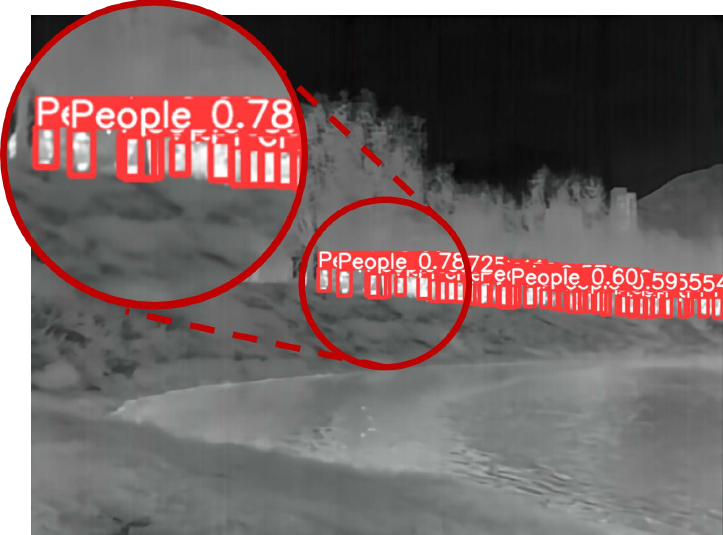}
		
		\\	
		\footnotesize	 UTV+LINF+IAT & \footnotesize UTV+KXNet+IAT & \footnotesize Ours
		\\
		\includegraphics[width=0.15\textwidth]{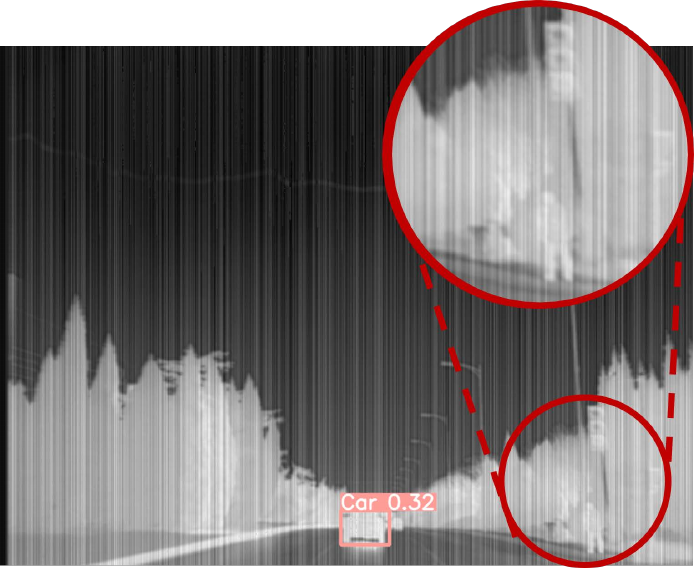}
		&\includegraphics[width=0.15\textwidth]{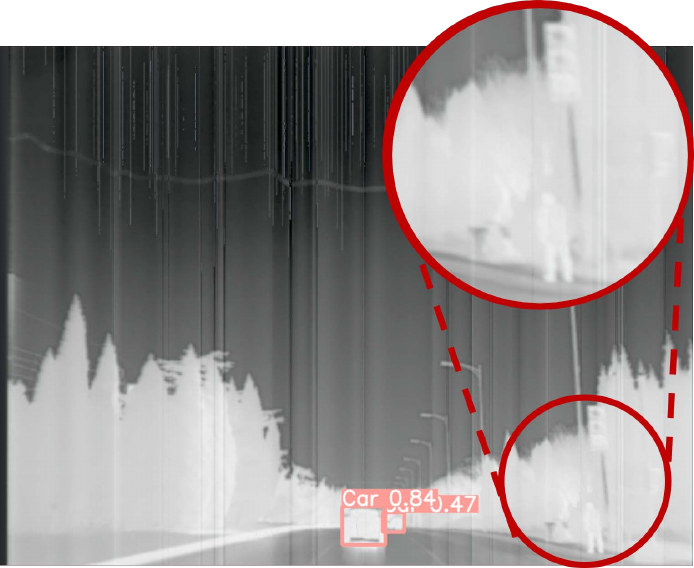}
		&\includegraphics[width=0.15\textwidth]{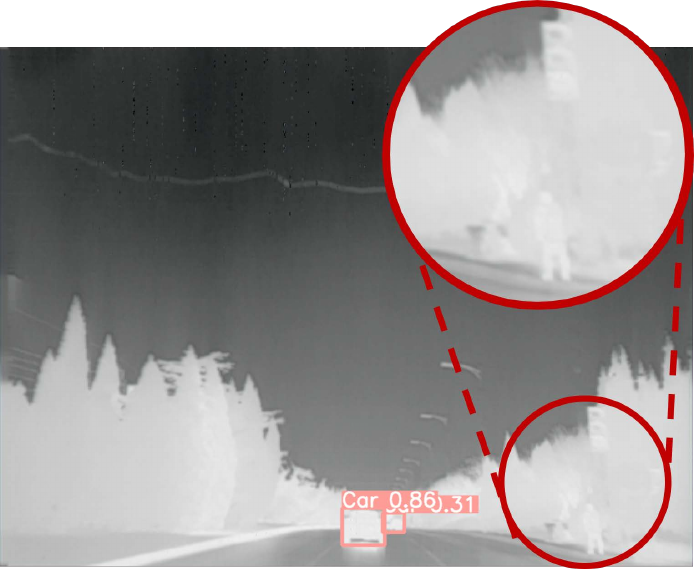}
		
		\\
		\footnotesize	 Degraded & \footnotesize SEID+LINF+IAT & \footnotesize SEID+KXNet+IAT
		\\
		\includegraphics[width=0.15\textwidth]{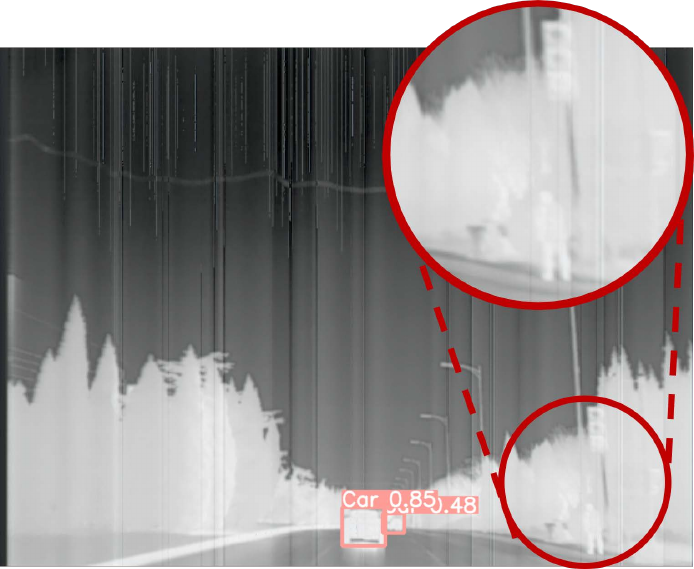}
		&\includegraphics[width=0.15\textwidth]{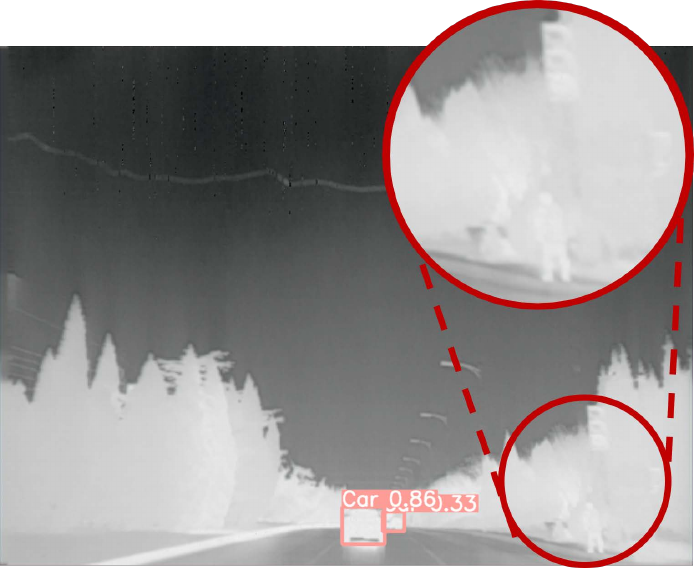}
		&\includegraphics[width=0.15\textwidth]{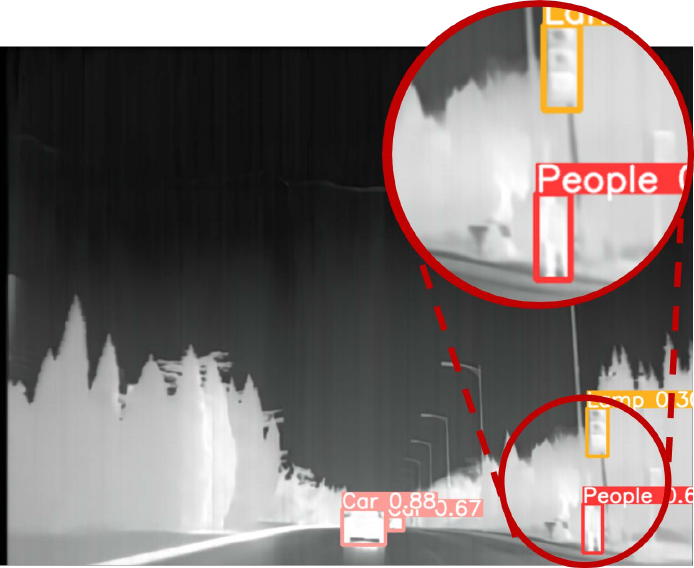}
		
		\\
		
		\footnotesize	 UTV+LINF+IAT & \footnotesize UTV+KXNet+IAT & \footnotesize Ours
		\\
	\end{tabular}
	\vspace{-1em}
	\caption{Object detection comparisons with several advanced competitors under three composited degradation factors.}
	\label{fig:detection}
\end{figure}

\section{Applications}

\begin{table}[tb]
	\centering
	\footnotesize
	\renewcommand{\arraystretch}{1.1}
	
	\setlength{\tabcolsep}{1.6mm}{
		\begin{tabular}{|c|ccccccc|}
			\hline
			\multicolumn{1}{|c|}{}        & \multicolumn{7}{c|}{Categories}                                                                                                                                                    \\ \hline
			\multicolumn{1}{|c|}{Methods} & \multicolumn{1}{c|}{People} & \multicolumn{1}{c|}{Car} & \multicolumn{1}{c|}{Bus} & \multicolumn{1}{c|}{Lamp} & \multicolumn{1}{c|}{Motor} & \multicolumn{1}{c|}{Truck} & mAP   \\ \hline
			\multicolumn{1}{|c|}{Degraded} & \multicolumn{1}{c|}{0.522} & \multicolumn{1}{c|}{0.345} & \multicolumn{1}{c|}{0.137} & \multicolumn{1}{c|}{0.161} & \multicolumn{1}{c|}{0.173} & \multicolumn{1}{c|}{0.233} & 0.262 \\ \hline
			S+K+I& \multicolumn{1}{c|}{\cellcolor{blue!10}{0.700}}& \multicolumn{1}{c|}{0.772} & \multicolumn{1}{c|}{\cellcolor{blue!10}{0.678}} & \multicolumn{1}{c|}{0.394} & \multicolumn{1}{c|}{0.444}& \multicolumn{1}{c|}{\cellcolor{blue!10}{0.683}} & \cellcolor{blue!10}{0.612} \\ \hline
			U+K+I& \multicolumn{1}{c|}{0.694}  & \multicolumn{1}{c|}{\cellcolor{blue!10}{0.808}} & \multicolumn{1}{c|}{0.667} & \multicolumn{1}{c|}{0.355} & \multicolumn{1}{c|}{0.480}       & \multicolumn{1}{c|}{0.640}  & 0.607 \\ \hline
			S+L+I                & \multicolumn{1}{c|}{0.695}  & \multicolumn{1}{c|}{0.771} & \multicolumn{1}{c|}{0.673} & \multicolumn{1}{c|}{\cellcolor{blue!10}{0.395}} & \multicolumn{1}{c|}{0.447}      & \multicolumn{1}{c|}{0.676} & 0.610 \\ \hline
			U+L+I                & \multicolumn{1}{c|}{0.695}  & \multicolumn{1}{c|}{0.806} & \multicolumn{1}{c|}{0.659} & \multicolumn{1}{c|}{0.358} & \multicolumn{1}{c|}{\cellcolor{blue!10}{0.481}}      & \multicolumn{1}{c|}{0.632} & 0.605 \\ \hline
			Ours                          & \multicolumn{1}{c|}{\cellcolor{red!15}{\textbf{0.737}}} & \multicolumn{1}{c|}{\cellcolor{red!15}{\textbf{0.826}}} & \multicolumn{1}{c|}{\cellcolor{red!15}{\textbf{0.726}}} & \multicolumn{1}{c|}{\cellcolor{red!15}{\textbf{0.436}}} & \multicolumn{1}{c|}{\cellcolor{red!15}{\textbf{0.528}}}      & \multicolumn{1}{c|}{\cellcolor{red!15}{\textbf{0.707}}} & \cellcolor{red!15}{\textbf{0.660}}  \\ \hline
		\end{tabular}
	}
	\vspace{-1em}\caption{ Numerical results of object detection compared with   four representative combinations on the M3FD dataset. }~\label{tab:detection}
\end{table}

\noindent\textbf{Depth estimation.}
We initially demonstrate the effectiveness of our method in the context of depth estimation using low-resolution infrared images. Fig.~\ref{fig:depth} illustrates the outcomes of depth estimation using the fundamental depth model, DepthAnything~\cite{yang2024depth} under three scenarios. It is notable that employing existing enhancement techniques fails to accurately predict depth for distant objects, as evidenced by the shapes of pedestrian and car. 
In contrast, our approach effectively highlights salient objects, significantly improving depth estimation accuracy.
%
%\textbf{Salient object detection.} Extracting typical objects from degraded scenarios poses significant challenges. In Fig.~\ref{fig:sod}, we evaluate the performance of our method under stripe noise interference using the classical salient object detector, U2Net~\cite{qin2020u2}. Results from MaskedDN and SEID are notably affected by streak noises, resulting in inaccurate object predictions. In contrast, our method effectively removes noise and enhances structures, allowing for precise estimation of distant cars in both scenarios. Additionally, the road scene in the third instance combines two complex degradations, including stripe noise and low contrast corruptions. Major objects such as power towers cannot be accurately estimated using other denoising-specific enhancement algorithms. In contrast, our approach considers diverse degradation factors comprehensively, enabling precise estimation of fine-grained structures in infrared images, highlighting both salient objects and structural details effectively.
%

\noindent\textbf{Object detection.} We demonstrate the compatibility of our approach for  object detection task based on YoLoV5~\cite{liu2022target}.  The numerical results are compared with various combined schemes in Table~\ref{tab:detection}. Our method achieves consistently optimal precision across all categories, demonstrating a significant improvement of almost 8\% in accuracy compared to the second-best method. Furthermore, we select two challenging scenes, one featuring small dense objects and another with a low-contrast scenario in Fig.~\ref{fig:detection}. The first scene includes tiny dense targets, combined with several challenging degradation factors. Existing methods are unable to accurately detect dense pedestrians due to blurred object structures resulting from the denoising procedure. In contrast, our method can successfully detect all small and dense objects. The second scene presents a challenging low-contrast scenario, where existing methods fail to effectively highlight targets and shapes, leading to detection failures. In contrast, our method can effectively detect all targets, including pedestrians and traffic lamps.

\section{Conclusion}
This paper proposed a data-efficient adversarial learning scheme to address dynamic complex corruptions in high-quality thermal imaging. Based on the dynamic adversarial solution, the degradation generation module effectively learns to simulate thermal-specific corruption, addressing the scarcity of high-quality thermal data. A dual interaction architecture was proposed, integrating scale transform and spiking network for thermal-specific features. 
Extensive experiments illustrated the superiority of our schemes.

{
	\small
	\bibliographystyle{ieeenat_fullname}
	\bibliography{reference}
}


\end{document}